# Do humans and Convolutional Neural Networks attend to similar areas during scene classification: Effects of task and image type

Romy Müller[1], Marcel Dürschmidt[1], Julian Ullrich[1,2,3], Carsten Knoll[2], Sascha Weber[1], & Steffen Seitz[2]

[1] Faculty of Psychology, Chair of Engineering Psychology and Applied Cognitive Research, TUD Dresden University of Technology, Dresden, Germany

[2] Faculty of Electrical and Computer Engineering, Chair of Fundamentals of Electrical Engineering, TUD Dresden University of Technology, Dresden, Germany

[3] Faculty of Mathematics and Natural Sciences, Department of Computer Science, Machine Learning Group, Heinrich-Heine-Universität Düsseldorf, Düsseldorf, Germany

* Corresponding author

E-mail: romy.mueller@tu-dresden.de (RM)

# Abstract

Deep Learning models like Convolutional Neural Networks (CNN) are powerful image classifiers, but what factors determine whether they attend to similar image areas as humans do? While previous studies have focused on technological factors, little is known about the role of factors that affect human attention. In the present study, we investigated how the tasks used to elicit human attention maps interact with image characteristics in modulating the similarity between humans and CNN. We varied the intentionality of human tasks, ranging from spontaneous gaze during categorization over intentional gaze-pointing up to manual area selection. Moreover, we varied the type of image to be categorized, using either singular, salient objects, indoor scenes consisting of object arrangements, or landscapes without distinct objects defining the category. The human attention maps generated in this way were compared to the CNN attention maps revealed by explainable artificial intelligence (Grad-CAM). The influence of human tasks strongly depended on image type: For objects, human manual selection produced maps that were most similar to CNN, while the specific eye movement task has little impact. For indoor scenes, spontaneous gaze produced the least similarity, while for landscapes, similarity was equally low across all human tasks. To better understand these results, we also compared the different human attention maps to each other. Our results highlight the importance of taking human factors into account when comparing the attention of humans and CNN.

*Keywords:* Convolutional Neural Networks (CNN), Explainable Artificial Intelligence (XAI), attention maps, eye movements, scene viewing, image classification, categorization

# Introduction

Human-technology cooperation could greatly benefit from recent advances in Deep Learning. For instance, Convolutional Neural Networks (CNN) are powerful image classifiers that sometimes match or even surpass human abilities (Buetti-Dinh et al., 2019). Modern CNN-based model structures such as ResNet (He et al., 2016) can support humans in potentially life-saving tasks such as medical image analysis (Kshatri & Singh, 2023) or scene understanding (Xie et al., 2017). Recent research additionally indicates that CNN-based networks achieve results comparable to state-of-the-art image processing models (Liu et al., 2022) such as Vision Transformers (Dosovitskiy et al., 2021) or Swin Transformers (Liu et al., 2021). However, CNN act like a black box, while successful human-technology cooperation requires transparency (Christoffersen & Woods, 2002; Klein et al., 2004). The transparency of CNN can be enhanced by methods of explainable artificial intelligence (XAI), which generate attention maps to highlight the image areas that contributed to the CNN's classification decision. Such transparency is crucial, as humans' trust in a CNN may deteriorate when it uses image areas that do not make sense to them – even when this does not result in inferior performance (Nourani et al., 2019). Accordingly, similarity to human attention has rightfully been suggested to be a relevant dimension for evaluating the quality of CNN and XAI methods. But how similar or different are the areas attended by CNN and humans, actually? Comparisons of XAI visualizations with human attention maps indicate that CNN sometimes base their classification on different image areas than humans (e.g., Rong et al., 2021; van Dyck et al., 2021; Zhang et al., 2019). However, it is unclear what factors are responsible for this. More precisely, most studies selectively focused on technological factors, such as the type of CNN or XAI. In contrast, factors that affect human attention have not been in focus, such as the procedure used to elicit human attention maps (e.g., eye tracking vs. manual selection) or the images to be classified (e.g., single objects vs. complex scenes). A striking diversity in these factors can be observed across studies. Although this is likely to affect the results of comparisons between humans and CNN, the specific impact of such human-centered factors is poorly understood, and thus will be addressed in the present study.

To facilitate reasoning about the similarity between human and CNN attention maps, we first provide a brief overview of the psychological literature on scene viewing. We ask what factors guide human eye movements in real-world scenes, and highlight some general differences to the image processing of CNN. After this, we review and integrate previous studies that compared human and CNN attention maps, focusing on the methodological diversity in tasks and images. Based on this overview, we derive our own experimental approach. Note that we may switch between the terms of scene "categorization" when talking about humans and "classification" when talking about CNN. This is to stay within the terminology of the respective literatures, but the terms should be understood as synonyms.

## Visual scene processing of humans and CNN

### How do humans process scenes?

Humans can infer the basic-level category of scenes at a glance. The representation of this so-called *gist* refers to the overall meaning of a scene, including perceptual as well as conceptual aspects (Oliva, 2005): people instantly extract not only low-level features (e.g., spatial frequencies, color) but also high-level semantic information (e.g., birthday party). Gist perception relies on two complementary information sources: global scene statistics and diagnostic objects. On the one hand, there are regularities in the structural and color patterns of scenes from different categories, which enable fast and reliable categorization (Torralba & Oliva, 2003). These physical scene statistics result in subjectively perceivable global properties (e.g., openness, temperature, or dynamics) that are vital for categorization (Greene & Oliva, 2009). For instance, deserts are high in openness and temperature but low in dynamics, while waterfalls are low in openness and temperature but high in dynamics. Accordingly, people are less likely to confuse deserts and waterfalls than deserts and wheat fields. On the other hand, diagnostic objects provide relevant information. Sometimes a single object is enough to infer the scene category (Wiesmann & Võ, 2023). For instance, people may be able to infer that a scene is an office merely based on the presence of a computer screen. Taken together, scene statistics and objects allow people to extract the gist of a scene in an instant.

Subsequently, this gist representation helps people decide where to move their eyes for more thorough analysis. Such eye movements are needed because people can only see sharply within the small area of foveal vision, and thus have to sequentially fixate relevant parts of the image (Findlay & Gilchrist, 2003; Holmqvist et al., 2011). Such *fixations target informative areas* which typically contain objects, while fixations of uniform background areas such as the sky or desert sand are rare (Henderson, 2003). Ample evidence suggests that it is the semantic relevance or meaningfulness of image areas that controls where people look, even when these areas are not physically salient (Henderson et al., 2009).

How do people know whether an area is meaningful, without already having looked at it? This is largely due to *scene context*: based on learned knowledge about statistical regularities, people can predict where meaningful objects are likely to be found (Henderson, 2017). Similar to initial gist processing, such contextual guidance of eye movements draws on the two complementary sources of global scene statistics and object-to-object relations. First, physical scene statistics and the resulting global properties determine where people move their eyes (Torralba et al., 2006). This is because objects are systematically organized along horizontal layers, so that people can expect airplanes to appear in the sky and pedestrians on the ground. A second important source of contextual guidance are object-to-object relations (Võ et al., 2019), because particular objects systematically co-occur in the real world. On the one hand, some objects serve as anchors for others (Boettcher et al., 2018). For instance, when looking for a laptop, people may initially fixate a table. On the other hand, subsequent fixations land on semantically related objects even when this relation is not hierarchical (Hwang et al., 2011). For instance, after fixating a plate, people are more likely to fixate a fork than a chandelier. Due to these systematic relations, people can also miss objects when their size does not match the rest of the scene, even when they are larger than normal (Eckstein et al., 2017).

The relative importance of global properties and object-to-object relations may depend on the *type of scene* (Wiesmann & Võ, 2023): global properties are particularly informative for outdoor scenes that differ in spatial layout, while objects are more informative for indoor scenes. The prominent role of transitions between semantically related objects may lead to a higher predictability of eye movements in indoor scenes. Conversely, eye movements are less deterministic for landscapes, which encourage more exploration (Wu et al., 2014). These findings indicate that it is worthwhile to consider the type of image to be categorized when comparing the attention maps of humans and CNN.

Although scene viewing is highly efficient, eye movements do not only depend on the current task. People are also prone to systematic *viewing biases*. One such source of distraction is the saliency of physical features (Itti & Koch, 2000). There is an ongoing debate whether the influence of saliency can fully be ascribed to meaning, as the two factors are highly correlated (Henderson et al., 2019; Pedziwiatr et al., 2022). However, for current purposes it suffices to note that sometimes task-irrelevant features can catch the human eye. Perhaps the most prominent example is social stimuli like faces, which reflexively capture attention (Rösler et al., 2017). Another task-independent influence on eye movements is central fixation bias (Tatler, 2007): people tend to look at the center of an image, even when the relevant contents are located in the periphery. To understand the effects of such systematic viewing tendencies on comparisons between human and CNN attention maps, we need to consider how CNN process scenes.

## How does CNN scene processing compare to that of humans?

CNN are specifically designed to process image data as they can take into account the relations between neighbouring pixels and are inspired by the visual information processing in biological brains. However, unlike humans, standard CNN models *do not have selective attention* but raster the entire image (but see Lin et al., 2016). This is important when interpreting the meaning of CNN attention maps as revealed by XAI methods: in contrast to human attention maps, non-highlighted areas in CNN maps do not tell us that the CNN has not thoroughly processed these areas, but merely that they did not contribute to the classification decision, despite having been processed just as much. For the sake of simplicity, we will still refer to both human and CNN maps as attention maps, but it should be noted that this does not mean the same thing in both cases. Moreover, CNN process the image information in several layers of neurons, and deeper networks with more layers result in higher representative capacities. Different layers process *different types of information* (Bau et al., 2017): whereas early layers focus on low-level features like colors and textures, later layers are responsible for high-level concepts like shapes and objects.

Given this general approach to image classification, what scene contents are used by CNN? In what ways does this resemble human scene processing and how does it differ? And where might CNN in fact have similar biases? As excellent discussions of this comparison can be found elsewhere (e.g., Firestone, 2020; Geirhos et al., 2020; van Dyck et al., 2021), we will selectively focus on the aspects discussed in the previous section on human scene viewing that are relevant to the present study. That is, we will emphasize the role of contextual guidance and its sources (i.e., global properties and objects) as well as the presence of task-irrelevant biases, while not considering other issues such as the role of image distortions or adversarial attacks.

A first dimension for comparison is the *role of scene context*. Similar to humans, CNN strongly rely on context. They tend to select classes that match this context, so that, for instance, snow may help CNN to classify an image of a dog as a husky (Ribeiro et al., 2016). However, humans usually benefit from compatible, typical context, but are still able to flawlessly categorize objects and scenes when the context is atypical. In contrast, atypical context affects the performance of CNN in remarkable ways and can lead to misclassifications (Beery et al., 2018; Geirhos et al., 2020): CNN may fail to recognize objects in unexpected locations (e.g., cows at the beach) and may classify non-existing objects when the context is suggestive (e.g., sheep when processing images of hills with green grass). This indicates that CNN sometimes use context much more than the actual objects to be classified. This heavy use of context mainly relies on structural scene statistics, while CNN have problems with the second form of contextual guidance, namely object-to-object relations. Accordingly, most computational approaches perform worse when classifying indoor scenes than landscapes (Quattoni & Torralba, 2009). An example of the restricted ability of CNN to deal with object relations pertains to relative object size. A positive effect is that, unlike humans, CNN do not miss targets when their size is unusual (Eckstein et al., 2017). However, the flip side is that CNN tend to mistake objects for visually similar ones (e.g., confusing brooms and toothbrushes), not taking into account that the object size is implausible given the other objects in the scene.

A second important dimension for comparison is the role of *task-irrelevant biases*. First, it seems like CNN attend to the saliency of image features in general (e.g., edges, luminance, or colors), not just the class-defining object (Zhang et al., 2019). However, the mechanisms behind this impact of saliency are likely to differ. For humans, the effects of saliency can largely be attributed to its high correlation with meaning (Henderson et al., 2019). It is unlikely that CNN also extract such meaning, and they may even rely on salient image areas that have nothing to do with the class as humans conceive of it. Thus, the areas attended by CNN may not make sense to humans (Meske & Bunde, 2020; Singh et al., 2020). Many of these divergences can be described to "excessive invariance" (Jacobsen et al., 2019): CNN learn whatever shortcut is sufficient for classification, and this may or may not correspond with the intended scene characteristics according to human standards (Geirhos et al., 2020).

Given these similarities and differences in scene processing between humans and CNN, it can be expected that the measurable outputs, namely attention maps, might also differ. The following sections will summarize the available research on comparing human and CNN attention maps, extract the factors that may affect this comparison, assess what previous studies have to say about these factors, and specify a research gap on which these studies remain silent.

## Comparing human and CNN attention maps

### Overview of findings

When comparing the attention maps of humans and CNN, a common finding is that their similarity is quite low (Das et al., 2017; Ebrahimpour et al., 2019; Hwu et al., 2021; Karargyris et al., 2021; van Dyck et al., 2021). Human attention tends to be more selective and focused on specific areas, while CNN attention is more diffuse and distributed (Hwu et al., 2021; Yang et al., 2022). Some studies found that CNN put more weight on context than humans. For instance, CNN attention maps may highlight the mere presence of body parts such as fingers or lips to classify skin diseases (Singh et al., 2020). A related finding is that the areas attended by humans are more discriminative and diagnostic (Rong et al., 2021). Accordingly, several studies have found CNN performance to improve when trained with human attention (Boyd et al., 2023; Karargyris et al., 2021; Lai et al., 2020). Aside from these general observations, the similarity between human and CNN attention maps is affected by *technological factors*. First, it may depend on the type of CNN (Lai et al., 2020; van Dyck et al., 2021; Zhang et al., 2019). For instance, similarity is higher for deeper networks with more layers (Zhang et al., 2019) or when CNN are designed to process information in a way that is more similar to humans, for instance via biologically plausible receptive field sizes (van Dyck et al., 2021) or human-inspired attention mechanisms (Das et al., 2017; Lanfredi et al., 2021). A second technological factor that affects similarity is the XAI method used to elicit attention maps (Ebrahimpour et al., 2019; Muddamsetty et al., 2021; Rong et al., 2021; Yang et al., 2022). For instance, some XAI methods focus on edges, while others use broad image areas, and some provide chunky and pixelated areas, while others provide smooth and gradual areas that resemble human attention distribution. In contrast to this focus on technological factors, previous studies have rarely investigated the effects of factors that affect human attention maps. In the present study, we investigate two of these factors: the tasks used to elicit human attention maps, and the images to be categorized. Before specifying our research questions, we will give an impression of the variability in these factors across previous studies.

## Influence of tasks

We use the term "task" to refer to the procedures of eliciting human attention maps. These procedures differ between studies in two nested ways: how directly they assess attention, and how this assessment is implemented in specific cognitive activities. Concerning the first distinction, a rather direct assessment approach is to track people's eye movements during scene categorization, while a less direct approach is to let people manually select the image areas they consider relevant. Within these two general approaches, previous studies used various cognitive activities, that is, different categorization procedures or different means of manual selection. The following section is organized by the general assessment approaches and then reviews the variety of specific activities within them.

*Eye movements*

Eye tracking is often considered the gold standard for eliciting human attention maps, and thus most studies have applied this method (Ebrahimpour et al., 2019; Hwu et al., 2021; Karargyris et al., 2021; Lanfredi et al., 2021; Muddamsetty et al., 2021; Rong et al., 2021; Schiller et al., 2020; Trokielewicz et al., 2019; van Dyck et al., 2021; Yang et al., 2022). In these studies, eye movements were tracked while people had to perform a wide variety of cognitive activities, which differed on several dimensions. One such dimension is the *amount of experimental control*. On the one hand, there have been rather unrestricted tasks such as free verbal descriptions during routine radiological image reading (Karargyris et al., 2021; Lanfredi et al., 2021), or driving in natural environments (Hwu et al., 2021). On the other hand, there have been highly controlled tasks such as performing saccades to briefly presented images (van Dyck et al., 2021). In between these extremes, various categorization procedures have been implemented, such as verbal labelling of images or image parts (Ebrahimpour et al., 2019; Schiller et al., 2020), keypress responses to choose between categories (Lai et al., 2020; Rong et al., 2021; Trokielewicz et al., 2019), or textual explanations of why the image matches a particular category label (Yang et al., 2022). A second dimension on which previous tasks differed is the degree to which they encouraged people to *focus their eye movements* on a specific area or to broadly explore the image. Some tasks required attention to small details, such as the fine-grained classification of birds (Lai et al., 2020; Rong et al., 2021) or the explicit description of features responsible for classification (Yang et al., 2022). Other tasks diverted eye movements more broadly across the image, such as naming as many objects as possible (Ebrahimpour et al., 2019) or providing a comprehensive report of radiological images (Karargyris et al., 2021; Lanfredi et al., 2021). A third dimension pertains to *restrictions of viewing time*. This ranged from extremely short viewing times (e.g., 150 ms), which only allows for one fixation (van Dyck et al., 2021), over medium viewing times (e.g., 3 s), which allows for a brief inspection (Ebrahimpour et al., 2019; Rong et al., 2021; Schiller et al., 2020) up to unlimited viewing time, which grants people the opportunity to thoroughly investigate the image (Hwu et al., 2021; Karargyris et al., 2021; Lanfredi et al., 2021; Trokielewicz et al., 2019; Yang et al., 2022). Obviously, tasks can be distinguished on many more dimensions (e.g., task difficulty or familiarity), but most of them are hard to evaluate for an outsider.

However, not only the variability of these tasks but also the *method of eye tracking per se* must be evaluated critically. First, eye movements convey information that is not relevant for scene categorization, due to systematic viewing biases. Second, they fail to convey other information that actually is relevant, because scene categorization can proceed in an instant, without any need for eye movements. Moreover, eye movements reflect people's search processes (instead of their decision, as it is the case for CNN), and thus fixations may land on areas where the target is likely to be (based on contextual constraints) but then turned out not to be present: good guesses that upon closer inspection turned out to be wrong (Henderson, 2017). These tendencies might call into question the suitability of eye movements to indicate which image areas people need for categorization. Thus, it is worthwhile to contrast them with other tasks to elicit attention maps, which do not suffer from the same problems.

*Manual selection*

Compared to eye tracking studies, the number of studies that used some form of manual selection to elicit human attention maps is much smaller (Das et al., 2017; Mohseni et al., 2021; Singh et al., 2020; Zhang et al., 2019). At the same time, their diversity is even larger. Manual selection studies differed in how people defined the areas relevant for categorization, and in the degrees of freedom they had. One type of task is to *define the outlines of relevant areas*, either by drawing polygons (Singh et al., 2020) or by lassoing them (Mohseni et al., 2021). Such procedures provide a binary value (0/1) for each participant and image, and gradual variations in attention maps only emerge from the aggregation over several participants. Conversely, another method can immediately provide gradual relevance estimates for each participant (Das et al., 2017): participants viewed a blurred image and had to *deblur relevant areas* by repetitive rubbing with the mouse cursor. This enabled a slight deblurring of areas that were only coarsely searched, and a complete deblurring of areas that upon closer inspection turned out to be actually relevant. For one, this procedure provides a nice analogy to eye movements, which also combine a quick ambient "where" processing of the layout with a subsequent focal "what" analysis of object details (Unema et al., 2005). Finally, one study elicited human attention maps by asking participants to *order pre-defined image segments* according to their relevance for classification (Zhang et al., 2019).

An important dimension that differentiates between these manual selection tasks is the likelihood of including scene context. First, including context can be encouraged by means of deblurring, where context processing is an integral part of the elicitation procedure (Das et al., 2017). Second, including context can be up to participants when they are free to select whatever areas they want (Singh et al., 2020). Third, it can be discouraged when segment ordering by relevance pushes participants to start with all parts belonging to the category-defining object (Zhang et al., 2019). Finally, it can be prevented entirely, when only the image parts inside an object segmentation mask are available for selection (Mohseni et al., 2021).

Aside from the specific implementation, *manual selection per se* comes with a number of advantages and disadvantages. These are complementary to those of eye tracking. Manually selecting relevant areas is a highly conscious, intentional activity, and may not adequately reflect the information humans actually use for categorization. On the one hand, they might conceive of their task as selecting the image areas that define an object, instead of the image context they actually need (Zhang et al., 2019). However, the other direction is also possible: humans may select more than they actually need for categorization, especially when the scene category is defined by large proportions of the image, as in the case of landscapes. Thus, they may select all areas of equal importance, even when in fact they need much less to reliably categorize the scene. Taken together, humans might not be well aware of their inner processes of categorization, and thus may not even know what information they are attending to.

*Comparing tasks*

Given the diversity of tasks implemented in previous studies, it is surprising that no attempts have been made to systematically investigate their impacts on the similarity between human and CNN attention maps. A few studies assessed the impacts of task characteristics on human attention maps per se, but unfortunately did not relate these results to CNN. First, Yang et al. (2022) performed a pilot study to compare two eye tracking procedures: free viewing versus explaining why an image matched its class label. Free viewing led to more distributed eye movements that often targeted irrelevant objects, while explanation led to eye movements that were more focused on key features. Second, Das et al. (2017) used progressive manual deblurring during visual question answering, and performed a pilot study to compare three task versions that varied whether people knew the original image and correct answer. A medium level of information (i.e., seeing only the blurred image but knowing the answer) produced attention maps that were most helpful for new participants in answering the questions. However, as no previous studies have assessed how task variations affect the similarity between humans and CNN, we can only speculate about that. Fig 1 presents a continuum of intentionality on which different elicitation tasks can be located. Eye movement task are placed on the low intentionality end of the continuum, whereas manual selection tasks are placed on the high end. Presumably, tasks at the low end reflect attentional processes more directly, but also produce more false positive information. Conversely, task at the high end can specify the category as imagined by humans, but also produce false negatives and redundant information.

**Fig 1. Continuum of intentionality on which different tasks to elicit human attention maps can be located.** The arrows represent different dimensions that can increase or decrease with intentionality. The direction and darkness of the arrows marks the direction of increase.

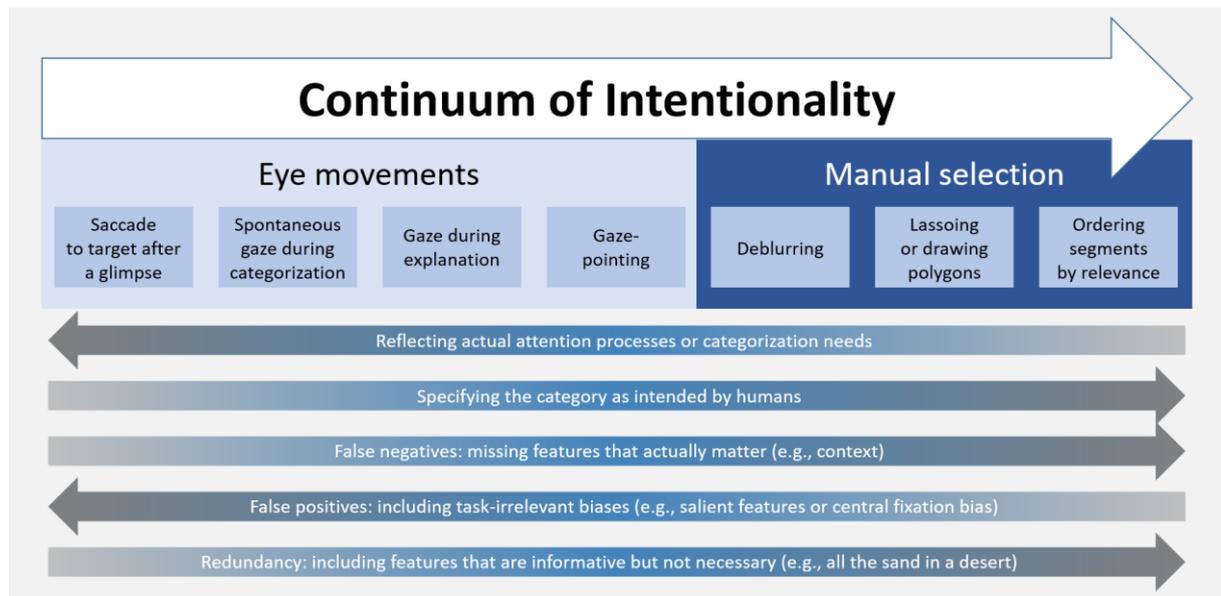

## Influence of image types

Similar to the tasks used to elicit attention maps, the type of images used for categorization showed considerable variation between studies. Dimensions of variation include image complexity, ambiguity of the areas relevant for classification, structural similarity of the images, and relevance of specific image details. The following section will discuss some of these differences and review studies that compared different image types more or less explicitly.

*Differences in image types*

A first dimension of variation is *image complexity*. Some studies used simple images with only one centrally presented, salient object (e.g., Mohseni et al., 2021; van Dyck et al., 2021), while others used complex scenes with multiple objects (e.g., Das et al., 2017; Ebrahimpour et al., 2019; Zhang et al., 2019). Again others used images that can only be interpreted by experts, such as medical images (Karargyris et al., 2021; Lanfredi et al., 2021; Muddamsetty et al., 2021; Singh et al., 2020; Trokielewicz et al., 2019). Second, images varied in the *ambiguity* of areas relevant for categorization. Presumably, it is rather straightforward to select relevant areas when the image only contains one salient object, which is likely to result in a higher similarity between humans and CNN. Conversely, it is more ambiguous which areas define landscapes. For instance, for deserts it seems impossible to unequivocally select the most relevant parts of the sand, and even other areas may be informative, such as the sun or clear blue sky. Sometimes image complexity and selection ambiguity may diverge, for instance when the category is clearly defined by a specific object in a highly complex scene (Das et al., 2017; Zhang et al., 2019). Third, the *similarity* of images within one and the same study varied. For instance, a high structural similarity is characteristic for human faces (Schiller et al., 2020) or medical images (Karargyris et al., 2021; Lai et al., 2020; Muddamsetty et al., 2021; Trokielewicz et al., 2019): while most of the image is identical in each trial, only specific details differentiate between the categories. In contrast, for natural scenes each image may have a completely different layout. A fourth, related dimension is whether the category distinction depends on a *small specific detail*, for instance during fine-grained classification of similar bird species (Lai et al., 2020; Rong et al., 2021). Such images do not need to be structurally similar, but still prompt humans to only focus on the most discriminative areas. Given this enormous variability in the images to be categorized, it is problematic that most previous studies did not make the characteristics of their images sufficiently explicit.

*Comparing image types*

Based on the scene viewing literature, it seems likely that the similarity of human and CNN attention maps depends on image type. A few isolated observations support this assumption, although most of them only reflect qualitative post hoc reports. On the one hand, influences of image type can emerge when some images are more prone than others to unintended shortcuts used by CNN. For instance, while classifying skin diseases, images with particular body parts yielded low similarity, because the CNN looked at the mere presence of lips, hair, or fingernails, instead of the actual skin condition (Singh et al., 2020). Moreover, higher similarity was observed when salient, discernible image areas were task-relevant. First, abnormal chest X-rays yielded higher similarity than normal ones, presumably because they contained areas of interest that attracted the attention of both humans and CNN (Lanfredi et al., 2021). Similarly, images of animate objects yielded higher similarity than images of inanimate objects, perhaps because they drew attention to faces (van Dyck et al., 2021). Taken together, task and image characteristics are likely to affect human attention maps and thereby change the similarity between humans and CNN. However, the current state of research only provides insufficient information about these influences.

# Present study

The present study investigated how the similarity between human and CNN attention maps depends on the task used to elicit human attention maps and the type of image to be categorized. To this end, we conducted an experiment in which humans had to assign scene images to one of six categories via keypress responses. In different parts of the experiment, we varied the intentionality of the task to elicit human attention maps, using two types of eye tracking tasks and a manual selection task. Within all three tasks, we varied the type of image, manipulating whether the categories mainly depended on objects, object-to-object relations, or global properties. We compared the resulting human attention maps to CNN attention maps generated by a common XAI method (Grad-CAM, Selvaraju et al., 2017). Henceforth, we will refer to our attention maps via the name of the specific elicitation method (e.g., spontaneous gaze, Grad-CAM) instead of via the agent whose attention is elicited (e.g., human, CNN) or the general elicitation approach (e.g., eye tracking, XAI). We aim to investigate how similarity depends on task characteristics, image characteristics, and their interaction.

## How does similarity depend on the task used to elicit human attention maps?

Our tasks represented three points on a continuum of intentionality (see Fig 1). On the low end, we simply tracked participants' *spontaneous gaze* during categorization. The aim of this task was to obtain fixations on areas that participants actually used, not confounded by participants being unaware of their true information needs. At the same time, this task comes with three risks. First, it might produce only few fixations, because eye movements are not needed for rapid gist perception. If participants do move their eyes, this leads to the second risk, namely that fixations might mainly reflect response selection processes (i.e., remembering the key mapping). In the best case, participants might look at the category-defining areas while selecting their response. In the worst case, it might lead to the third risk, namely that eye movements reflect task-irrelevant biases. In sum, attention maps generated by spontaneous gaze might provide little information about the areas actually relevant for categorization.

Given these risks, attention maps can be elicited by moving to the other end of the continuum of intentionality: manual selection. This was implemented in a task we will call *drawing*: participants used their mouse to draw a polygon around the most relevant area. This avoids the risks of spontaneous gaze, but comes with the risk of participants not being aware of their inner processes. This could take two forms. First, participants might select the whole area that defines a category (e.g., all the sand in a desert). While this choice is valid given that natural scenes are defined by global properties, it might include areas that people would never actually attend to. Second, the opposite risk is that participants might select an arbitrary part of the scene (e.g., a small patch of desert sand). Presumably, this area would be different for each participant, providing little generalizable information.

Considering the complementary risks of spontaneous gaze and drawing, we introduced a third task to combine the benefits and mitigate the costs: *gaze-pointing*. We instructed participants to intentionally fixate the areas most relevant for categorization. This task was inspired by two previous findings. First, tracking participants' eyes while they explained why an image matched its category label led them to focus on key features (Yang et al., 2022). Second, gaze-pointing caused eye movements to be more focused on relevant parts of a scene than free viewing (Müller et al., 2009). Gaze-pointing still is likely to include fixations on task-irrelevant areas, but might compensate for this by putting additional weight on task-relevant areas that participants would not spontaneously look at (e.g., desert sand).

From a practical perspective, gaze-pointing can tell us whether attention maps benefit from additional instruction or spontaneous gaze is sufficient to elicit useful attention maps.

Concerning the similarity of human attention maps to CNN, different outcomes are conceivable. On the one hand, if our concerns about spontaneous gaze and drawing are warranted, this should lead to higher similarity with gaze-pointing than the two other tasks. On the other hand, if the concerns about one or both of these tasks are unwarranted, we might find different results, depending on which task actually is problematic. However, we also expected these effects to be highly dependent on image type, which is why we consider the interaction to be most informative.

## How does similarity depend on the type of image to be categorized?

Our image types aimed to capture relevant distinctions from the psychological literature on scene viewing. For each image type, we used two separate but similar categories in order to make our task sufficiently difficult. Our first image type, which we will refer to as *objects* (i.e., lighthouse, windmill), only required the identification of a single diagnostic object to infer the category. This object was embedded in a natural scene context, which certainly facilitates categorization, but is not strictly necessary. Our second image type comprised two *indoor scenes* (i.e., office, dining room). Here, the category can be inferred from several diagnostic objects and their object-to-object relations, whereas global properties are similar for the two categories. We intentionally selected two categories that include chairs and tables, which typically are the most salient objects in indoor scenes (Torralba et al., 2006). Finally, our third image type was *landscapes* (i.e., desert, wheat field), with the scene category mainly depending on global properties and the distribution of spatial frequencies, but not on diagnostic objects or their relations.

We hypothesized that the similarity between human and CNN attention maps would be largest for objects, as they provide a single key feature. Furthermore, we expected medium similarity for indoor scenes, due to a strong guidance of eye movements by semantic relations between objects. Finally, we expected the lowest similarity for landscapes, assuming eye movements to be widely distributed across the image. However, as indicated in the previous section, we were most interested in the interaction between task and image type.

## How do task and image type interact?

We hypothesized that task influences would strongly depend on image type. First, for objects we expected human-CNN similarity to be consistently high across all three tasks, with no clear differences between them. As the categories were defined by a locally restricted, salient and meaningful object, we assumed this object to attract human attention, both by focusing their gaze and constraining their manual selections. Thus, the risks of using eye movements should be negligible (i.e., low influence of task-irrelevant biases), and the relevant areas should be easy to select manually (i.e., simply drawing a polygon around the object).

Second, for landscapes we expected human-CNN similarity to be low across all three tasks, again with no differences between them. As you probably cannot mark specific areas in any coherent manner for large, uniform areas of sand or grain, this should lead to a high variability of human attention maps in all tasks. Thus, not even the more intentional forms of selection (i.e., gaze-pointing, drawing) are likely to produce consistent results. Drawings might fall between two strategies, with some participants selecting large proportions of the scene, some selecting arbitrary patches, and some selecting anything in between. For the two eye movement tasks, we expected participants to widely spread their gaze across the image. However, due to the powerful influence of scene guidance, eye movements were expected to be somewhat more systematic than drawings, perhaps even leading to higher similarity with CNN, which also rely global scene statistics.

Finally, we did expect task differences for indoor scenes. Due to the risks described above, spontaneous gaze and drawing might be less similar to CNN, while gaze-pointing might lead to comparably high similarity. This is because we expected indoor scenes to result in a pattern somewhere in between objects and landscapes. Just like landscape images, they might make it hard to select relevant areas, because relevance depends on large parts of the scene. Just like object images, object-based scene guidance might direct participants' gaze to particular indoor scene objects. However, the benefit should be higher for gaze-pointing, as it is assumed to compensate for systematic viewing biases. Taken together, we expected the influence of task to be strongest for indoor scenes.

# Materials and Methods

## Data availability

All images, human participant data, and source code (CNN, XAI, attention maps, comparison metrics) are made available via the Open Science Framework (https://osf.io/k9t5f/). Within this repository, the minimal dataset is to be found here: https://osf.io/rtf3u. The source code is additionally available on GitHub (https://github.com/cknoll/Humans-vs.-CNN-Effects-of-task-and-image-type).

## Participants

Twenty-eight members of the TUD Dresden University of Technology participant pool (ORSEE, Greiner, 2015) took part in the experiment between 22 September 2022 and 13 October 2022 in exchange for course credit or a payment of 8 € per hour. Due to occasional hardware problems, the eye tracker computer failed to store the eye movement files of three participants. Thus, the final sample consisted of 25 participants (16 female, 9 male) with an age range of 20 to 64 years ($M$ = 32.4, $SD$ = 10.9). Only participants who were fluent in German and had normal vision were included. The research was approved by the Ethics Committee at the TUD Dresden University of Technology (file sign: SR-EK-400092020), participants provided written informed consent, and all procedures followed the principles of the Declaration of Helsinki.

## Apparatus and stimuli

### Lab setup and eye tracking

The experiment took place in a lab room at TUD. Eye movements were tracked monocularly at 1000 Hz using the EyeLink 1000 infrared eye tracking system (SR Research Ltd., Ontario, Canada) with a chin rest and a viewing distance of 93 cm. Stimuli were presented on a 24" LCD display with a resolution of 1920 by 1080 pixels at a refresh rate of 60 Hz. A Cedrus Pad was used for keypress responses and a standard computer mouse was used to draw polygons around relevant image areas.

## Images

Images were taken from the Places365 dataset (Zhou et al., 2017), which provides a wide variety of images from 365 different classes. We chose this dataset because it offers a wide variety of different scene types (i.e., objects, indoor scenes, landscapes) compared to the heavily object-focused ImageNet dataset (Deng et al., 2009). The complexity of its natural scenes was important to us for three reasons. First, we needed a dataset that allowed us to investigate the distinction between images that rely on diagnostic objects versus object-to-object-relations versus global properties. Second, we wanted to increase the likelihood of being able to obtain eye movements at all, which would have been even more uncertain if we had used simpler images. Third, Places365 provides images at a much higher and consistent spatial resolution than other datasets like ImageNet, and in this way enabled us to present images in a format that is large enough to observe subtle differences in fixation locations.

We used the whole dataset for training the CNN, but only a subset of 102 images for the human experiment. Suitable images were selected, center-cropped, resized to a square format of 1024 x 1024 pixels. These images were then presented at the center of the screen in front of a white background. We manually went through the images to ensure that the relevant areas still were in full view, and replaced images where this was not the case. In total, 102 images were selected, 60 of them for the main experiment and 42 for practice.

For the main experiment, we selected 20 images for each of our three image types. Within each image type, we used two categories, and thus each category consisted of 10 exemplars (see Fig 2 for examples). To make categorization more challenging, these two categories were highly similar in terms of their scene statistics and global properties (e.g., openness, temperature). Other than that, we aimed for a high variability in the images for exploratory purposes. Accordingly, images differed in whether they included salient, category-irrelevant objects that might attract eye movements, and had no fixed ratio of the area covered by potentially category-defining versus less relevant contents.

For the image type *objects*, all scenes included one clearly discernible object of the respective category (i.e., lighthouse, windmill), and thus categories were unambiguously defined by local object information. While the objects were embedded in natural contexts, these contexts could be more or less informative. For instance, only some of our lighthouses were presented in front of a coastline. Some images also contained other, irrelevant objects (e.g., ships or cars). For *indoor scenes*, the images presented an arrangement of objects in a room with a specific function (i.e., office, dining room). All images included chairs and tables, and three offices also included a person. For *landscapes*, nature scenes with large, relatively uniform areas were used (i.e., desert, wheat field). Some landscapes also contained salient objects (e.g., agricultural machinery, houses).

Besides these images, the following additional stimulus screens were used in the experiment. First, at the start of the experiment participants saw a screen on which they had to input their demographic data (i.e., age and gender). Second, before each block, instruction screens summarized the respective task. However, the main instruction was provided in a video before the experiment. Third, in the first practice block, verbal category labels (instead of images) were presented centrally on a white background in black font (Tahoma, 30 pt). Finally, in the practice blocks a feedback screen informed participants about the correctness of their response. In case of an error, a schematic image of the Cedrus pad was shown that linked the category labels to the respective keys to remind participants of the correct key assignment. All materials were provided in German language.

**Fig 2. Stimulus examples for all three image types and the two corresponding categories.**

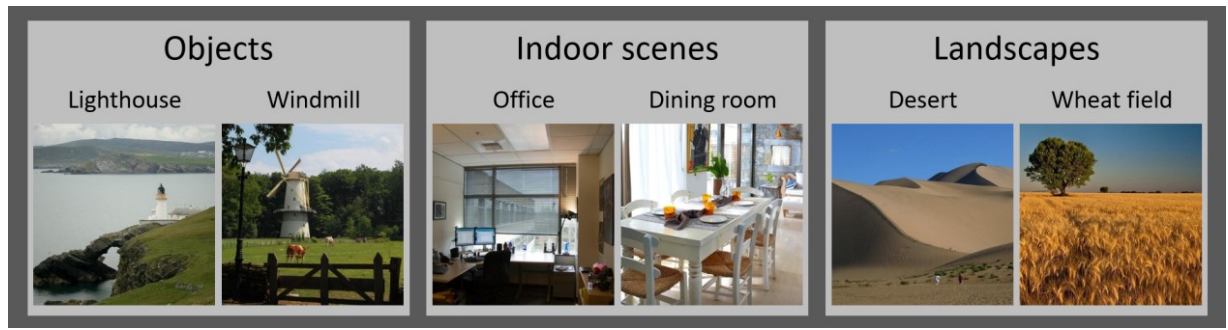

## Procedure

Throughout the experiment, participants had to assign images to six categories via keypress responses. The specifics of this procedure depended on the respective task (see below). In a within-participants design, we varied the two factors *task (spontaneous gaze, gaze-pointing, drawing)* and *image type (objects, indoor scenes, landscapes)*. The tasks varied between consecutive blocks in a fixed order, and image type varied randomly between trials. The same 60 images were used in all three blocks.

A session started with participants receiving a brief summary of the procedure and providing informed consent. They were then shown the instruction video and the eye tracker was calibrated. This was followed by two practice blocks (42 trials each), in which participants learned the key mapping and received feedback on the correctness of their responses. During the first practice block, they had to categorize words that corresponded to one of the six categories, and in the second practice block they had to categorize images. They also received feedback about their response and in case of an error, they were reminded of the correct key assignment. An overview of main experiment's procedure is provided in Fig 3. It consisted of three blocks (60 trials each, corresponding to the 60 images presented in random order). The blocks corresponded to the three tasks and always appeared in the same order. In the first block (i.e., spontaneous gaze), participants merely had to categorize the image by pressing a key. In the second block (i.e., gaze-pointing), participants had to categorize the image as well, but were asked to intentionally look at those areas of the image that were most relevant for their decision. In the third block (i.e., drawing), participants had to first categorize the image again, and after their keypress they had to draw a polygon around the relevant area with their mouse. Each mouse click defined a polygon point and after setting the last point, participants could press the lower left key of the Cedrus pad to connect the last point to the first one and thereby close the polygon.

The basic procedure of a trial was identical in each block, with the exception of the additional drawing procedure in the last block. A trial started with a drift correction, and participants had to fixate it while pressing the lower left key of the Cedrus pad to proceed to the image screen. The image then remained visible until participants pressed a key to indicate their category choice. The key assignment was randomly determined for each participant, and participants were asked to keep their middle fingers, index fingers, and thumbs on the six keys. In the main experiment, participants no longer received correctness feedback. Taken together, the experiment took about 45 minutes.

**Fig 3. Procedure of the main experiment.**

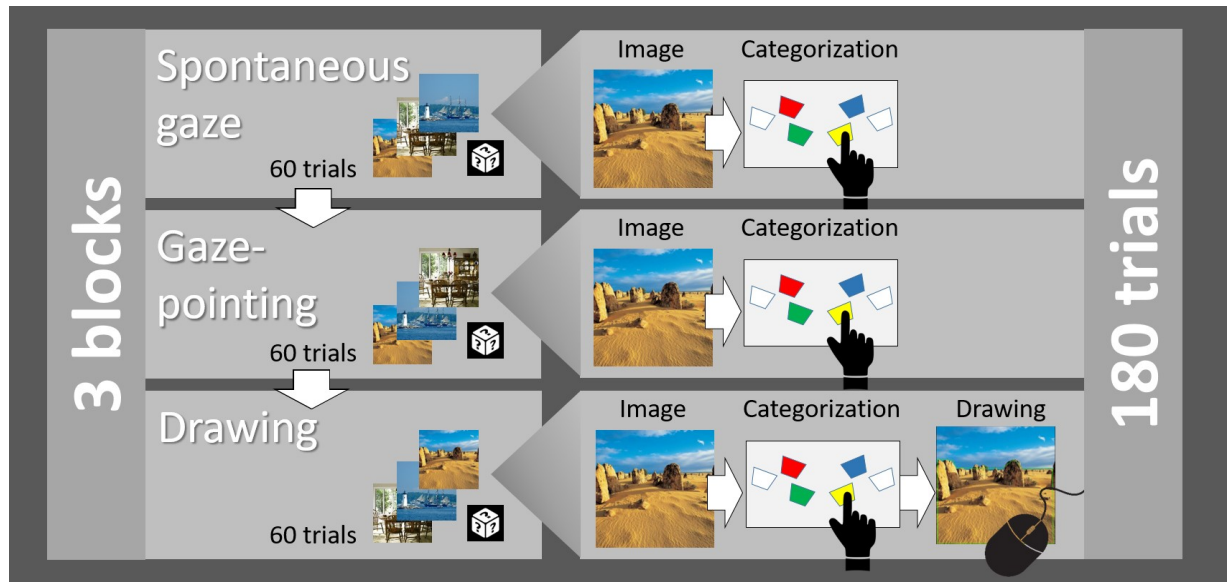

## CNN and XAI

### CNN model

The CNN model used in the present study was a ResNet-152 (He et al., 2016). This network was chosen because it is one of the architectures most frequently used the XAI literature, and particularly in previous investigations of human-CNN similarity (Das et al., 2017; Lai et al., 2020; Muddamsetty et al., 2021; Rong et al., 2021; van Dyck et al., 2021; Yang et al., 2022; Zhang et al., 2019), allowing for meaningful comparisons to previous research. While other models like Vision Transformers (a more complex neural network structure compared to CNN) are a promising approach for image classification, they currently lack support for XAI methods. Also, networks based on the ResNet architecture remain popular and are regularly used in contemporary state-of-the-art publications (Kolesnikov et al., 2020; Liu et al., 2022). The ResNet-152 consists of 152 consecutive layers, which are connected in a special structure. The basic idea is to iteratively apply convolutional filters to perform feature extraction and thus increase the information density in every layer. This is mostly done using bottleneck blocks, which consist of three convolutional layers (see Supporting Information, S1 Fig). The input and output of each block is connected via so-called residual connections. These residual connections are used to address the vanishing gradient problem, which can disrupt the training process for CNN that consist of many layers.

The input consists of 224 x 224 x 3 tensors (i.e., three-dimensional matrices), representing the pixel values in the RGB image. Through the application of the convolutional layers, the image is successively transformed into a collection of *feature maps*, which are matrices of activation values. The output of the convolutional part are 2048 feature maps of size 14 x 14, which are condensed to 2048 single scalar activation values via average pooling. On these values, a fully connected layer with linear weights is applied. This so-called classification head outputs a scalar score for each of the 365 classes. Finally, a softmax function is used which assigns a value between 0 and 100 to each class based on the respective score. This allows for an interpretation of the output as a percentage of the CNN's certainty that the image belongs to a particular class. The overall structure has approximately $58 \times 10^6$ free parameters (i.e., weights), which allows for a relatively fast training compared to more sophisticated CNN.

The network was trained with all images of the dataset (i.e., all 365 classes) for 10 epochs with the goal of minimizing the classification error, and it achieved a Top5-accuracy of 85%. The loss and accuracy curves are shown in the Supporting Information (S2 Fig). The training followed the standard procedure described by He et al. (2016): The weights are initialized with random values. Subsequently, images with known class values are fed into the net and the difference between the actual and the desired output is backpropagated to the weights, so that the weights are changed according to a cost function. This optimization process for the weights is repeated until overall classification performance converges. The problem of overfitting the CNN to the training images is addressed by validating the performance on test images which are not used in the training process. More details regarding the model and the training process can be found in the Supporting Information (S1 Table) and the model source code in our OSF repository.

To apply this ResNet structure to our 60 selected images with a resolution of 1024 x 1024 pixels, these images were downsampled to 224 x 224 in order to match the dimension of the input layer. Note that for the training process, a total of 1.825 million non-square images with varying resolution from the Places365 dataset were used. To make these images compatible with the input layer, transformations such as resizing and random-cropping were applied, as described by He et al. (2016).

## XAI method

The XAI method used in the present study was Grad-CAM (Selvaraju et al., 2017), which is the method used most often in previous comparisons of humans and CNN. The acronym expands to "*Gradient-weighted Class Activation Mapping*". Grad-CAM is a model-agnostic method, meaning that it is applicable to a wide range of CNN architectures without requiring adaptations of the internal structure. Furthermore, it is relatively straightforward to implement and results in low execution times. Technical details are provided in the section on attention map generation. Using spatial information obtained from the last convolutional layer, Grad-CAM yields an attention map, highlighting important areas of the input image. For the present study, highlights were generated for the respective target class, not the class that was deemed most likely by the CNN. For instance, in case the CNN misclassified a lighthouse for an oil rig, we still used the highlights for lighthouse.

## Data analysis

### Attention map generation

To generate human attention maps, the data (i.e., fixations or polygons) were summed over all 25 participants prior to map generation. We compared two types of attention maps between humans and CNN: binary masks of a fixed size, and gradual density maps. To define our binary masks, the same basic approach was used for all four types of attention maps. That is, we used the gradual density maps as a basis and set a cut-off that only kept those 5 % of the area visible that received the most weight, while the rest was hidden (see Fig 5 for examples of overlaying these masks onto images). In this way, all attention maps had the same size, while only their shape and position varied. The threshold of 5 % was chosen for two reasons. First, previous work has shown that areas as small as 5 % best differentiate between human and CNN attention, whereas for larger areas of about 20%, maps get much less distinguishable (Rong et al., 2021). Second, for object images (i.e., lighthouses, windmills), human eye movements were usually restricted to a rather small area, despite being summed over all participants, with area sizes ranging from 5.3 to 17.6 %, and an average of 11.0 %. Thus, using a higher threshold than 5 % would have required uncovering areas for some images that were never selected by any participant. However, to better understand how broadly participants actually spread their attention across the image in different task and image conditions, we additionally compared the sizes and variability of maps that uncovered all areas ever attended. The following sections will describe the specifics of area definition for different types of attention maps.

*Eye movement attention maps*

To generate our eye movement attention maps, we excluded all fixations outside the image, the first fixation (i.e., the one that started during drift correction), fixations with durations of less than 180 ms (Velichkovsky et al., 2005), and all fixations from trials with response times that deviated more than 2.5 SD from the average response time of 2106 ms (i.e., longer than 7629 ms). For the remaining data, the following procedure was used to generate fixation maps (for similar approaches see Lanfredi et al., 2021; Rong et al., 2021). Around each fixation, we considered an area of 2 degrees of visual angle (58 pixels), which corresponds to foveal vision. Within this area, we applied a Gaussian kernel that caused the weight to decrease from the center to the outside. This kernel was scaled so that the weight was 1 at the center of the fixation, and 1/58 at a distance of 58 pixels from the center (i.e., linear distance-weighting). Furthermore, each kernel was multiplicatively weighted by the fixation duration measured in milliseconds. The final attention map was generated by adding up the weighted kernels of all fixations. The resulting map was then normalized to the interval [0, 1] to simplify visualization. Note that we decided not to normalize the data for individual participants. That is, the areas only depended on fixation duration, so that participants with more or longer fixations had higher impact. This is because we saw no theoretical reason to discount the fixations of participants who scanned the image more thoroughly. However, we also performed exploratory analyses using a normalized procedure (i.e., each participant contributing the same weight) but found that this did not have any noteworthy impacts on the results.

*Drawing attention maps*

When defining the attended areas based on human drawings, we noticed some missing data due to technical problems (i.e., polygon coordinates were not recorded). This happened 20 times in total and thus removed 1.3 % of the data from the analysis. Other than that, we did not exclude any drawings, regardless of how many polygon points participants defined or how long it took them. To obtain the attention maps, we started with a matrix of the same resolution as the image with all values set to 0. If an area was inside the drawn polygon for a participant, the value in the matrix was increased by 1. Adding up all polygons then produced our attention map. Unlike the other attention maps, this procedure resulted in maps with hard edges, making it unclear which pixels should be used if we wanted to limit the mask size to exactly 5 % of the image. The final attention map for the drawings was thus obtained after smoothing the prior map. This was done via average pooling using a 3 x 3 kernel.

*Grad-CAM attention maps*

To generate CNN attention maps, Grad-CAM uses the gradients of any target class, flowing into the final convolutional layer to produce a coarse localization map that highlights those areas in the image that were important for predicting the class. For a given CNN and a selected class, the Grad-CAM algorithm generates an attention map, a so-called class-discriminative localization map (CDLM) for each input image. This is done by examining the activation flow from the last convolutional layer to the output (in other words: how the activation values influence the numerical score of the selected class). The result of the last convolutional layer can be interpreted as a collection of K feature maps, where in our case K = 2048. The last convolutional layer is chosen as it is expected "to have the best compromise between high-level semantics and detailed spatial information" (Selvaraju et al., 2017, p.4). In a first step, the gradient (i.e., the relative change of the result when changing the input) of the selected class with respect to each "pixel" in each feature map is calculated. These matrices are then averaged over the pixel dimension to obtain K so-called "importance weights" alpha_k. In a second step, each feature map is weighted by these alpha_k. Any negative values in the maps are set to 0 (via so-called ReLU nonlinearity), focusing the map only on areas with a positive impact on the class decision. This results in K weighted feature maps, which can be represented as an F x F x K tensor, were F denotes the feature map resolution. As a final step, the pixel-wise average of those maps is taken (i.e., along the last axis of the F x F x K tensor) to obtain the desired CDLM. Naturally, the CDLM has the same resolution (F x F) as the output of the last convolutional layer, which is typically much lower than that of the original image, in our case 14 x 14. To apply the CDLM to our input images, bilinear upsampling was used, resulting in an attention map with the same resolution as the original image (in our case 1024 x 1024). Such an attention map contains relevance values between 0 and 1 for every pixel in the original image, and could be visualized as a heatmap or density map, but it can also be transformed to a binary map. This was achieved by choosing the threshold for the relevance score such that only 5 % of the pixels (and thus of the total area) were included.

## Similarity calculation

We used two metrics to compute the similarity between attention maps: Dice score and cross-correlation. The Dice score (Dice, 1945) was used for binary masks, and specifies the overlapping area relative to the total attended area. It is calculated by taking two times the area highlighted in both maps to be compared (intersection) and dividing it by the sum of the two individual areas. If the areas have the same size (as it is the case in the present study), the Dice score simplifies to the size of the intersection divided by our chosen area size of 5 %. This procedure creates values between 0 and 1, with 0 indicating no overlap and 1 indicating complete overlap. This metric has already been used in previous studies to compare human and CNN attention (Schiller et al., 2020; Singh et al., 2020). It is not only easy to compute but also easy to interpret, because the numerical value directly corresponds to the share of overlapping area. As all maps are reduced to the same size, it also allows for a straightforward comparison of the overlap in different conditions (e.g., tasks or image types), even when the total attended areas systematically differ between them. However, this simplicity can be considered a limitation as well, because comparing binary masks eliminates the rich information available in gradual density maps. Therefore, we additionally compared the gradual maps using cross-correlation (Pearson), which has been deemed one of the most suitable metrics for purposes similar to ours (Bylinskii et al., 2019). Pearson's Correlation Coefficient treats two given density maps as variables and describes their linear correlation. We calculated the value for two density maps as proposed by Bylinskii et al. (Bylinskii et al., 2019) by dividing the covariance matrix for both maps by the product of the covariance of each map itself.

## Statistical analyses

Our statistical analyses aimed to compare human-CNN similarity in attention maps between tasks and image types. To this end, we conducted F2 analyses of variance (ANOVAs) with the 60 images as degrees of freedom (20 per image type), instead of using participants as degrees of freedom (F1 ANOVA). This is because the definition of human attention maps made it necessary to sum all fixations and drawings over participants, instead of using the areas attended by individual participants. An added benefit of these F2 ANOVAs is that we did not have to average across images but could consider the variance between individual images in our analyses. For ANOVA outcomes, we report the following values: (1) The F-statistic, which corresponds to the ratio of variation between sample means and variation within the samples (i.e., factor variance divided by error variance), (2) the p value, which indicates whether a difference is statistically significant, with p values < .05 reflecting significance, (3) partial eta squared ($\eta p^2$), which is a measure of the effect size, calculated as the proportion of total variance that is explained by the factor or interaction of factors after excluding variance from other factors. All pairwise comparisons were performed with Bonferroni correction. If the sphericity assumption was violated, a Greenhouse-Geisser correction was applied and the degrees of freedom were adjusted accordingly.

To analyze human-CNN similarity, we performed a mixed-measures F2 ANOVA with the three-level within-images factor *human-CNN comparison* (spontaneous gaze vs. Grad-CAM, gaze-pointing vs. Grad-CAM, drawing vs. Grad-CAM) and the three-level between-images factor *image type* (objects, indoor scenes, landscapes). To better understand the results of this human-CNN comparison, we used the same statistical approach to compare the three types of human attention maps to each other. That is, we replaced the factor *human-CNN comparison* with the factor *human-human comparison* (spontaneous gaze vs. gaze-pointing, spontaneous gaze vs. drawing, gaze-pointing vs. drawing). Moreover, we compared the total sizes of the areas attended in the three human tasks (i.e., without reducing them to 5 %) via a mixed-measures F2 ANOVA with the three-level within-images factor *task* (spontaneous gaze, gaze-pointing, drawing) and the three-level between-images factor *image type* (objects, indoor scenes, landscapes). Finally, we performed two control analyses, one that only used the first fixation in the two eye movement tasks, and one that split the three-level factor image type into its six constituent image categories. However, for the sake of brevity, we will not report these analyses in detail but only consider their results in the Discussion.

# Results

To support a better understanding of our human-CNN comparison results, we first report a number of qualitative observations. We then turn to the statistical analyses that compare the Dice score and cross-correlation of human and CNN maps between tasks and image types. After this, we explore potential reasons for these results by assessing how the human attention maps differed from each other. To this end, we first analyzed their Dice score and cross-correlation, and then examined differences in the size of the total areas uncovered. All mean values and standard deviations for the human-CNN comparisons as well as the human-human comparisons of attention maps are provided in Table 1.

## Qualitative observations

When visually inspecting the human attention maps and their overlaps with Grad-CAM, a number of noteworthy differences became apparent. Examples for some of the points are provided in Fig 4. As the comparison between attention maps was highly dependent on image type, we will structure the following section accordingly.

For *objects*, eye movements reflected several phenomena known from the scene viewing literature. For instance, fixations did not only target the category-defining object, but also other salient objects (e.g., boats in lighthouse images). Occasionally, Grad-CAM fell victim to the same biases. This typically happened when the context was atypical (e.g., for a lighthouse in an urban area, Grad-CAM focused on a car) but not when it was typical (e.g., for lighthouses by the sea, Grad-CAM did not look at boats). Thus, the resulting low overlap often stemmed from problems of gaze, not Grad-CAM. Accordingly, Grad-CAM overlapped more strongly with drawings, which did not suffer from these biases. However, Grad-CAM also had its problem with object images. First, it tended to look at the lower part of the lighthouse or windmill. In contrast, human eye movements and drawings preferably targeted the upper, more diagnostic part. Second, Grad-CAM even missed the object entirely when the context was highly atypical (e.g., lighthouse in the snow). Finally, human drawings also showed a typical pattern for objects: they did not include much context but focused on the object, sometimes aiming to precisely draw its outlines (e.g., individual rotor leafs of windmills). Thus, participants mainly varied in how carefully they specified the object boundaries, but largely agreed in selecting only the object.

For *indoor scenes*, we observed similar distraction in eye movements. For instance, fixations always landed on people when they were present in offices, while Grad-CAM only looked at a human face once. However, participants also included people in their drawings, suggesting that they actually considered them relevant for categorization. For drawings, participants generally used different strategies. Some selected an individual object (e.g., computer screen), some selected anchor objects (e.g., desk area), and some included almost the entire room. However, this also depended on the specific image category: including the entire scene was more common for offices than dining rooms, where people usually selected the table and the objects on it.

For *landscapes*, the similarity between eye movements and Grad-CAM was lowest. On the one hand, a few factors were conducive to similarity. For instance, both eye movements and Grad-CAM were sensitive to physically salient features (e.g., object boundaries, horizon), while rarely looking at non-informative areas (e.g., sand). Moreover, both tended to look at objects (e.g., agricultural machinery). On the other hand, the factors that reduced similarity played a larger role. For instance, fixations were biased towards the center even when it was non-informative, while Grad-CAM showed no signs of center bias. Moreover, Grad-CAM usually highlighted one area, while eye movements were widely dispersed across the scene (with the peaks of density maps appearing as blobs in the binary masks). For drawing, there seemed to be two major strategies, with some participants selecting most of the category-defining area (e.g., all desert sand) and some selecting an arbitrary part (e.g., a small patch of sand). However, many participants adopted strategies in between. Similar to indoor scenes, strategies differed between the two categories, with more variation for deserts than wheat fields.

Concerning the *comparison between human tasks*, we were quite surprised that the two types of eye movement maps rarely differed from each other, or only for individual images. For instance, when irrelevant objects were present, they had a stronger impact on spontaneous gaze, while gaze-pointing seemed to intentionally target non-salient but category-defining areas. While the eye movement maps were highly similar to each other, both their shapes and dispersion were quite different from drawing, but this difference seemed restricted to indoor scenes and landscapes.

**Fig 4. Example image overlays with attention maps to illustrate our qualitative observations.**

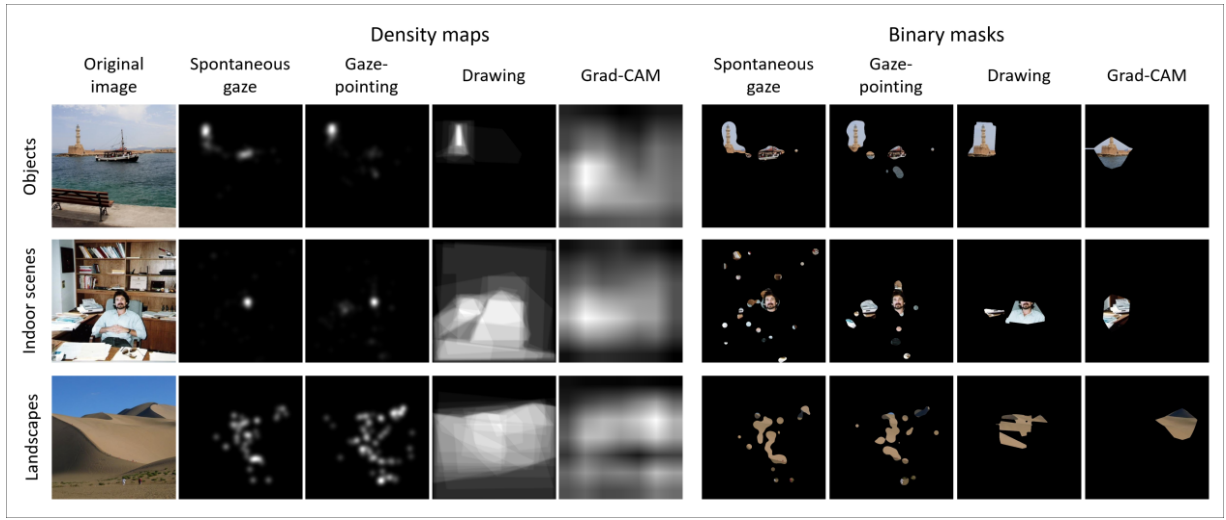

## Similarity between human and CNN attention maps

### Dice score

For the Dice score, the 3 (*human-CNN comparison: spontaneous gaze vs. Grad-CAM, gaze-pointing vs. Grad-CAM, drawing vs. Grad-CAM*) x 3 (*image type: objects, indoor scenes, landscapes*) ANOVA revealed a main effect of human-CNN comparison, $F(1.4, 80.9) = 6.438$, $p = .006$, $\eta_p^2 = .101$, a main effect of image type, $F(2, 57) = 23.558$, $p < .001$, $\eta_p^2 = .453$, and an interaction between the two factors, $F(4, 114) = 10.484$, $p < .001$, $\eta_p^2 = .269$ (see Fig 5A, red bars). The main effect of human-CNN comparison indicated that the human attention maps that overlapped most with Grad-CAM were those elicited by drawing. That is, Grad-CAM had a higher overlap with drawing than with spontaneous gaze (.30 vs. .24, respectively), $p = .019$, while comparing the overlap between Grad-CAM and drawing to that between Grad-CAM and gaze-pointing (.26) just missed significance, $p = .050$. Conversely, the two types of eye movement maps did not differ in their overlap with Grad-CAM, $p = .683$. The main effect of image type indicated that the overlap between human attention maps and Grad-CAM was highest for objects (.43), followed by indoor scenes (.26) and landscapes (.11), all $p$s < .005. Finally, the interaction revealed that the differences between human-CNN comparisons strongly depended on image type. For objects, the overlap with Grad-CAM was most dependent on human task. Here, Grad-CAM overlapped more with drawing (.53) than with spontaneous gaze and gaze-pointing (.40 and .37, respectively), both $p$s < .002. Conversely, the two eye movement tasks showed similar overlap with Grad-CAM, $p = .179$. For indoor scenes, spontaneous gaze overlapped least with Grad-CAM (.20), and this overlap was lower than for gaze-pointing and drawing (.28 and .31, respectively), both $p$s < .011. Finally, for landscapes, all three human tasks showed very little overlap with Grad-CAM (with .13, .13, and .06 for spontaneous gaze, gaze-pointing, and drawing, respectively), and no differences between them were found, all $p$s > .188.

### Cross-correlation

For cross-correlations, the ANOVA revealed a main effect of human-CNN comparison, $F(1.2, 69.2) = 122.978$, $p < .001$, $\eta_p^2 = .683$, but no main effect of image type, $F(2, 57) = 1.235$, $p = .298$, $\eta_p^2 = .042$, and no interaction, $F(4, 114) = 439$, $p = .780$, $\eta_p^2 = .015$ (see Fig 5B, red bars). The main effect of human-CNN comparison indicated that Grad-CAM correlated higher with drawing (.54) than with spontaneous gaze and gaze-pointing (.28 and .29, respectively), both $p$s < .001, while the two types of eye movement maps did not differ in their correlation with Grad-CAM, $p$ > .9. As indicated by the lack of an interaction, the same pattern was found for all three image types: higher correlations of Grad-CAM with drawing than with the two eye movement tasks, all $p$s < .001, and similar, low correlations of Grad-CAM with the eye movement tasks, all $p$s > .7.

## Human attention maps

### Dice score

Before considering the statistical analysis, an inspection of Fig 5A reveals that the overlaps between human tasks (grey bars, $M$ = .45) were consistently higher than the overlaps between humans and CNN (red bars, $M$ = .27). This was the case even for landscapes, which had produced very low human-CNN overlap, whereas the two eye movement tasks still overlapped more with each other than Grad-CAM had overlapped with any eye movement task for any image type. The 3 (*human-human comparison: spontaneous gaze vs. gaze-pointing, spontaneous gaze vs. drawing, gaze-pointing vs. drawing*) x 3 (*image type: objects, indoor scenes, landscapes*) ANOVA yielded a main effect of human-human comparison, $F(1.8, 103.0) = 50.139$, $p < .001$, $\eta p^2 = .468$, a main effect of image type, $F(2,57) = 54.290$, $p < .001$, $\eta p^2 = .656$, and an interaction, $F(4,114) = 4.318$, $p = .004$, $\eta p^2 = .132$. The main effect of human-human comparison indicated that all comparisons between human tasks yielded different degrees of overlap, all $p$s < .001. That is, the two eye movement tasks had the highest overlap (.55), followed by gaze-pointing vs. drawing (.44), and then spontaneous gaze vs. drawing (.37). The main effect of image type indicated that the overlap between human attention maps was highest for objects (.61), followed by indoor scenes (.44), and landscapes (.30), all $p$s < .005. Finally, the interaction indicated that the differences between human-human comparisons depended on image type. However, the direction of these dependencies was opposite to what we had observed for the human-CNN comparison: for objects, the overlap was *least* (instead of most) dependent on human tasks, because all of them were highly similar. Accordingly, the only difference that just passed the significance threshold indicated that spontaneous gaze overlapped more with gaze-pointing (.66) than with drawing (.57), $p = .034$. No other differences were found, all $p$s > .2. For indoor scenes, the two eye movement tasks still had a high overlap with each other (.54), while the overlap between drawing and either spontaneous gaze or gaze pointing (.36 and .42, respectively) was considerably lower, both $p$s < .002. These two latter comparisons did not differ from each other, $p = .100$, indicating that drawing maps were generally quite different from eye movement maps for indoor scenes. Finally, for landscapes, all comparisons between human tasks were significant, all $p$s < .001, with the highest overlap between the two eye movement tasks (.44), medium overlap between drawing and gaze-pointing (.30) and very low overlap between drawing and spontaneous gaze (.17).

### Cross-correlation

The ANOVA revealed a main effect of human-human comparison, $F(1.6,90.7) = 227.835$, $p < .001$, $\eta p^2 = .800$, a main effect of image type, $F(2,57) = 7.632$, $p = .001$, $\eta p^2 = .211$, but no interaction, $F(4,114) = .760$, $p = .554$, $\eta p^2 = .026$ (see Fig 5A, grey bars). The main effect of human-human comparison indicated that all comparisons between human tasks yielded different correlations, all $p$s < .001. While the two eye movement tasks had the highest overlap (.73), the correlation between drawing and gaze-pointing (.45) also was higher than that between drawing and spontaneous gaze (.39). The main effect of image type indicated that correlations were lower for objects (.43) than indoor scenes (.61), *p* = .001, while landscapes (.54) yielded correlations that were not significantly different from either of the two other image types, both *p*s > .084. The absence of an interaction indicated that a similar pattern of human-human comparisons was found for all image types, with the two eye movement tasks correlating higher with each other than with drawing, all *p*s < .001. Moreover, drawing showed slightly higher correlations with gaze-pointing than with spontaneous gaze for both objects and landscapes, both *p*s < .035, but not for indoor scenes, *p* = .431.

**Table 1. Means and standard deviations (in parentheses) for all human-CNN comparisons and human-human comparisons, depending on task and image type.**

|  |  | Dice score | | | Cross-correlation | | |
| --- | --- | --- | --- | --- | --- | --- | --- |
|  |  | Objects | Indoor scenes | Landscapes | Objects | Indoor scenes | Landscapes |
| Human-CNN | SponG_Grad-CAM | .40 (.17) | .20 (.16) | .13 (.10) | .31 (.06) | .24 (.13) | .30 (.07) |
|  | GPoint_Grad-CAM | .37 (.17) | .28 (.17) | .13 (.17) | .33 (.08) | .26 (.14) | .28 (.09) |
|  | Draw_Grad-CAM | .53 (.19) | .31 (.27) | .06 (.11) | .55 (.21) | .52 (.21) | .55 (.19) |
| Human-human | SponG_GPoint | .66 (.08) | .54 (.11) | .44 (.09) | .64 (.09) | .80 (.16) | .74 (.20) |
|  | SponG_Draw | .57 (.09) | .36 (.17) | .17 (.14) | .28 (.08) | .50 (.22) | .40 (.24) |
|  | GPoint_Draw | .60 (.07) | .42 (.16) | .30 (.16) | .38 (.06) | .53 (.15) | .46 (.15) |

SponG = spontaneous gaze, GPoint = gaze-pointing, Draw = drawing.

**Fig 5. Comparisons between attention maps depending on image type, for both human-CNN comparisons (red bars) and human-human comparisons (grey bars). (A) Dice score and (B) Cross-correlation.** SponG = spontaneous gaze, GPoint = gaze-pointing, Draw = drawing. Error bars represent standard errors of the mean.

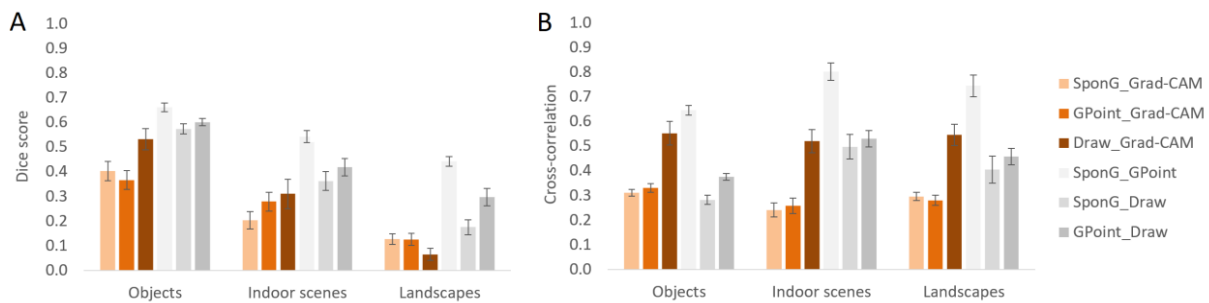

## Size of attended areas

Did different tasks lead participants to select smaller or larger image areas, and how strongly did this vary between image types and individual images? We compared the total attended areas on each image, summed over all participants (i.e., all areas that were ever selected by any participant). Examples of the smallest and largest attention maps are presented in Fig 6, and an overview of the respective means, standard deviations, minima, and maxima is provided in Table 2. The 3 (*task:*

*spontaneous gaze, gaze-pointing, drawing*) x 3 (*image type: objects, indoor scenes, landscapes*) revealed a main effect of task, $F(1.1, 62.8) = 272.536$, $p < .001$, $\eta_p^2 = .827$, a main effect of image type, $F(2,57) = 127.075$, $p < .001$, $\eta_p^2 = .817$, and an interaction, $F(4,114) = 58.741$, $p < .001$, $\eta_p^2 = .673$ (see Table 2). The main effect of task indicated that in total, larger areas were selected via drawing (49.5 %) than either spontaneous gaze or gaze-pointing (17.4 and 18.7 %, respectively), both *p*s < .001, while the two eye movement tasks did not differ significantly, *p* = .066. The main effect of image type indicated that the areas for objects (11.5 %) were smaller than for indoor scenes and landscapes (37.7 and 36.4 %), both *p*s < .001, while the two scene-centric image types did not differ, *p* > .9 Finally, the interaction indicated that the task-dependence of area sizes strongly varied with image type. For objects, spontaneous gaze, gaze-pointing and drawing did not differ (11.0, 9.1 and 14.3 %, respectively), all *p*s > .190. For indoor scenes, much larger areas were selected via drawing (73.7 %) than either spontaneous gaze or gaze-pointing (18.5 and 20.9 %, respectively), both *p*s < .001, while the two eye movement tasks did not differ significantly, *p* = .069. For landscapes, the areas again were much larger for drawing (60.5 %) than the two eye movement tasks, *p* < .001, but this time also larger for gaze-pointing (26.2 %) than spontaneous gaze (22.6 %), *p* = .002.

**Table 2. Area sizes as percentage of the total image area, depending on task and image type.**

|  | Objects | | | | Indoor scenes | | | | Landscapes | | | |
| --- | --- | --- | --- | --- | --- | --- | --- | --- | --- | --- | --- | --- |
|  | M | SD | Min | Max | M | SD | Min | Max | M | SD | Min | Max |
| SponG | 11.0 | 3.0 | 6.3 | 17.6 | 18.5 | 3.5 | 12.2 | 24.3 | 22.6 | 6.8 | 13.1 | 35.2 |
| GPoint | 9.1 | 2.7 | 5.3 | 17.6 | 20.9 | 5.0 | 13.3 | 29.8 | 26.2 | 4.8 | 18.7 | 36.1 |
| Draw | 14.3 | 7.2 | 4.9 | 29.5 | 73.7 | 15.8 | 46.9 | 96.8 | 60.5 | 17.9 | 30.9 | 93.6 |

SponG = spontaneous gaze, GPoint = gaze-pointing, Draw = drawing, M = mean, SD = standard deviation, Min = minimum, Max = maximum.

**Fig 6. Area sizes and their variability.** The rows represent the smallest (first row) and largest (third row) area size for the respective combination of task and image type, a typical area (second row) represents the area size closest to the mean of the respective combination.

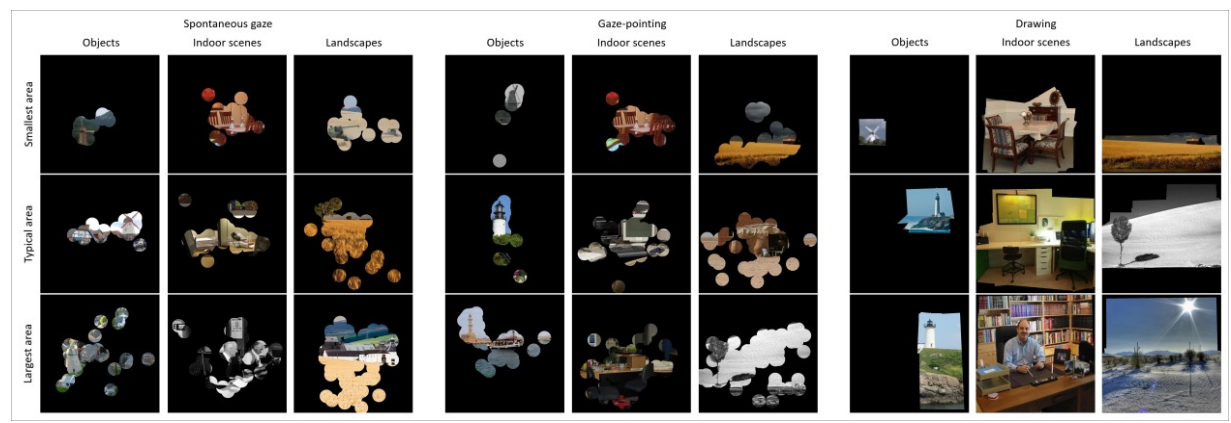

# Discussion

Do humans and CNN attend to similar areas during scene classification? And how does this depend on the task used to elicit human attention maps, the type of image to be classified, and the interaction between tasks and images? To answer these questions, we varied the intentionality of human image area selection, comparing two eye movement tasks (i.e., spontaneous gaze, gaze-pointing) and manually selection (i.e., drawing). Participants performed these tasks on three types of images that reflected important determinants of human scene viewing: the category either relied on a single diagnostic object, on object-to-object relations, or on global scene properties (i.e., objects, indoor scenes, and landscapes, respectively). We compared the resulting attention maps to those generated by a common XAI method (i.e., Grad-CAM). This comparison was realized by computing two similarity metrics (i.e., Dice score, cross-correlation). These metrics produced consistent results concerning the question how human-CNN similarity depended on the human task, but led to diverging conclusions about whether this interacted with image type (which it did according to Dice score but not according to cross-correlation). The following discussion will first focus on the results obtained via the Dice score that compares which areas received most attention, but we will turn to the differences between the two metrics below.

## Overview of results

Human attention maps were much less similar to CNN attention than they were between different human tasks. This fits with the widely reported finding that the similarity between human and CNN attention maps is quite low (e.g., Das et al., 2017; Ebrahimpour et al., 2019; Hwu et al., 2021; Karargyris et al., 2021; van Dyck et al., 2021). It also fits with a specific finding that inter-human similarity is higher than human-CNN similarity (Das et al., 2017). Thus, in general, our findings are in line with previous research. Moreover, human-CNN similarity depended on both our human-centred factors. Concerning task, attention maps derived from drawing were more similar to Grad-CAM than those derived from eye movements, while the specific eye movement task had little impact. Concerning image type, the similarity in the areas most attended by humans and CNN was highest for objects, medium for indoor scenes, and low for landscapes. While we cannot relate the task-dependent differences to previous human-CNN comparisons (because task influences have not been investigated so far), the image-dependent differences match previous findings. Specifically, higher similarity between humans and CNN has been found for abnormal X-rays with a clearly discernible relevant area (Lanfredi et al., 2021) and for animate objects where faces presumably caught human attention (van Dyck et al., 2021). Thus, the main attentional focus of humans and CNN is more similar when the relevant area is non-ambiguous and spatially restricted.

Our most interesting findings concern the interaction between both factors, with the direction of task effects being reversed depending on image type: for objects, drawing generated much higher similarity to Grad-CAM than the two eye movement tasks, while for landscapes, drawing was descriptively least similar to Grad-CAM, although the difference between tasks was not significant here. Thus, the results that can be obtained when eliciting human attention maps with manual selection are highly dependent on image type, at least when comparing where most attention was focused. This might lead to the conclusion that manual selection only is suitable when there is an obvious, correct solution (i.e., specific object), but not when selection is up to human preferences. It needs to be noted, though, that the suitability of an elicitation task cannot be inferred from high similarity values per se. The latter would be circular reasoning when the aim is to *test* similarity rather than *generate* it. However, suitability can be inferred from the variability and thus potential arbitrariness of attention maps, which also was much higher for drawing than eye movements for indoor scenes and landscapes. We will discuss this issue in more detail below.

## Why was manual selection more similar to CNN than eye movements?

Overall, manual selection generated attention maps that were most similar to CNN. Although no previous study has compared manual and gaze-based elicitation of attention maps, this result seems to be at odds with a previous study (Zhang et al., 2019). This study attributed the dissimilarity between manually generated human attention maps and CNN to the fact that relevant scene context was considered by CNN, but neglected by humans. Based on these results, we had hypothesized that eye movement maps (at least for gaze-pointing) would be more similar to CNN than manually generated maps, because eye movement control is guided by scene context (Henderson, 2017; Torralba et al., 2006; Võ et al., 2019). However, this hypothesis was not supported by our data, and at least for objects the opposite was found. To account for this discrepancy, it needs to be noted that a coarse representation of scene gist can be established without eye movements (Oliva, 2005), and relevant objects already are targeted by the first or second fixation (Henderson et al., 2009). Thus, our prediction might have been somewhat naïve: only because context guides eye movements, this does not necessarily mean that people actually look at this context. Even if they do, we probably would not have captured this in our data, given that we excluded fixations shorter than 180 ms, because these ambient fixations reflect orientation in space rather than identification (Velichkovsky et al., 2005).

A second reason for the higher similarity between drawings and Grad-CAM could be that neither of them was prone to task-irrelevant biases such as central fixation bias or attentional capture by salient distractors. In contrast, these biases were found in eye movement maps, decreasing their overlap with Grad-CAM. A third reason might be our specific method of manual selection. Participants could freely select image areas by drawing polygons around them, while Zhang et al. (2019) had participants order small, pre-defined image segments by relevance. Accordingly, our participants did include considerable amounts of context in their drawings, which becomes apparent when analyzing the size of selected areas: manually selected areas were as large as those selected by eye movements for objects, and much larger for indoor scenes and landscapes.

However, we also observed considerable variance in individual participants' manual selections for indoor scenes and landscapes (see Fig 4), suggesting that the areas participants select for these image types can be somewhat arbitrary. This is in line with previous research. First, Das et al. (2017) found that although inter-human similarity in manual selections was higher than human-CNN similarity, it still was numerically quite low. Second, Mohseni et al. (2021) did not explicitly report inter-human similarity, but their image examples of superimposed manual selections from 10 participants show considerable differences. Thus, it seems like a high variance in manual selections is not unusual.

## Why were the two eye movement tasks so similar to each other?

Another noteworthy finding was that the two eye movement tasks produced highly similar attention maps, and thus rarely differed in their similarity to Grad-CAM. Only for indoor scenes, gaze-pointing overlapped more with Grad-CAM than spontaneous gaze, and even this was only true for the main focus of attention compared via Dice score. Two explanations might account for this difference. First, when categorization depends on an arrangement of different objects, participants instructed to look at relevant areas might intentionally fixate different objects, even though a single diagnostic object would be sufficient for categorization (Wiesmann & Võ, 2023). Second, gaze-pointing might lead participants to intentionally fixate objects that are more diagnostic but less salient (e.g., not only fixating the person in an office but pointing out a printer).

Aside from this small difference for indoor scenes, the eye movement tasks produced highly similar results. Conversely, previous research had shown that eye movements during free viewing were more dispersed than eye movements recorded while participants intentionally performed gaze-pointing (Müller et al., 2009) or explained the suitability of category labels (Yang et al., 2022). To account for this difference, it should be noted that in these previous studies, the task goals differed more strongly than they did in our study: free viewing is not identical with spontaneous gaze during categorization. The former allows for inspecting whatever image areas may seem interesting, whereas spontaneous gaze during categorization is likely to target category-defining areas, which participants would also point out intentionally. Given that people have to solve a specific task, they already look at task-relevant areas from the first fixation onwards (Henderson et al., 2009). Such early influences of task-relevance also were corroborated by our control analysis that only used the first fixation to generate attention maps: the overlap between spontaneous gaze and gaze-pointing was not affected much, and numerically it even increased from .55 to .58.

An additional explanation for the high similarity between our eye movement tasks is based on systematic tendencies in scene viewing. For one, both eye movement tasks were affected by task-irrelevant biases and salient objects. However, the high similarity was also observed for landscapes, where eye movement maps had an overlap of .44, which is higher than most human-CNN comparisons (except drawing for objects). This is noteworthy given that landscapes encourage eye movements to be quite exploratory (Wu et al., 2014), which was reflected in a high dispersion of fixations and a corresponding blobbiness of attention maps (see Fig 4). However, even the small blobs often coincided between the two tasks. This highlights the important role of scene guidance (Henderson, 2017; Torralba et al., 2006; Võ et al., 2019). From a practical perspective, it suggests that when aiming to elicit consistent attention maps for images without clearly identifiable relevant areas, eye tracking might be a better choice than manual selection.

When reasoning about the choice of a particular eye movement task, it should be recalled that we had originally included gaze-pointing as a last resort: an alternative task that could be used if spontaneous gaze only produced non-interpretable results, due to the automaticity of gist perception or the pertinence of biases. However, given the high similarity of the two eye movement tasks, this concern seems unwarranted: researchers can simply use spontaneous gaze during categorization, as it rarely differs from more intentional selection.

## Why did the two similarity metrics produce different outcomes?

When comparing the results obtained with Dice score and cross-correlation, the overall effects of tasks were consistent, whereas their dependence on image type was strikingly different. Only for the Dice score, task effects on human-CNN similarity varied with image type: Grad-CAM was more similar to drawing than to the eye movement tasks for objects, but these task differences were largely diminished for indoor scenes and completely disappeared for landscapes. In contrast, cross-correlations did not depend on image type: Grad-CAM was always more similar to drawing than eye movements.

How can this divergence between the two metrics be explained? At first glance, one might assume that the cross-correlation results are an artefact of total area size, because in contrast to the Dice score, cross-correlations do not control for task differences in area size. These area sizes were large for Grad-CAM and drawing, but small for the two eye movement tasks. Specifically, the total eye movement areas never exceeded 37 % even for indoor scenes and landscapes (see maximum values in Table 2). This means that more than 60 % in each image remained completely unattended (black parts of the density maps in Fig 4). Conversely, in the drawing task such unattended areas only amounted to about 3 and 6 % for indoor scenes and landscapes, respectively. Grad-CAM also assigned at least some relevance to all image areas. This should result in a higher correlation between drawing and Grad-CAM even for indoor scenes and landscapes. Thus, the explanation would be that cross-correlations increase with a broad focus of attention (or lack of specificity), and therefore remain high for drawing, while the eye movement tasks fostered more specificity (cf. Rong et al., 2021) and thus showed lower correlations with the non-specific Grad-CAM.

However, this explanation of cross-correlation results being a mere artefact of non-specificity is refuted by another observation: the cross-correlations between the human tasks. While the total area sizes became more different between drawing and eye movements for indoor scenes and landscapes (as compared to objects), the cross-correlations between them *increased* rather than decreased. Thus, differences in attention breadth do not automatically reduce cross-correlations.

Therefore, instead of arguing which metric is more valid, we should consider that they answer different questions. On the one hand, the Dice score can be used if you want to compare which areas humans and CNN attend to most. In this case, image type has a major impact on the results. On the other hand, cross-correlation can be used if you want to compare how humans and CNN generally deploy their attention across the entire image. In this case, image type does not seem to matter. That said, it should be noted that cross-correlation has the advantage of retaining more variance and thus being able to detect the nuances, instead of forcing to select a specific area while the actual attention deployment is not in line with such specificity for some image types. What remains consistent across both metrics is the finding that overall, the areas attended by Grad-CAM are more similar to those elicited with drawing than eye movements.

## Particularities of task implementation

When discussing task influences on human-CNN similarity, we need to consider a number of problems with our tasks. A first problem is that they were quite arbitrary in some implementation details that are likely to affect the results. For instance, if we had used fewer or more categories, or categories that were more or less similar, this might have changed which areas participants attended. If participants had needed to choose between only two categories, they would probably have made very few fixations, because response selection would have been much easier. Conversely, if they had needed to choose between 20 or 200 categories, our procedure would not have been feasible at all, because participants would have been unable to memorize the key mapping. Thus, our tasks do not easily scale to more complex categorization demands. Some of these implementation issues could be solved by minor changes to the procedure, such as performing verbal instead of manual labelling (e.g., Schiller et al., 2020) or presenting a category label beforehand and then having participants indicate whether it matches the image or not (e.g., Biederman et al., 1982).

The available category alternatives can also affect eye movements in other ways. Given the obvious differences between our image types, it presumably was sufficient to scan the image superficially: even just looking at a person in an office will tell you that this is not a desert or lighthouse, and probably also not a dining room. Conversely, if there had been several similar categories (e.g., office, home office, reception desk, computer lab, computer store), participants would have been forced to fixate the most informative, discriminative areas. A similar point has been made for human-CNN comparisons that relied on fine-grained classification (Rong et al., 2021), but in these images, only a singular feature differentiated between the categories. Thus, an interesting question for future research is how humans and CNN differ in discriminating complex scenes that consist of similar objects. It is questionable, however, whether this can be tested with spatial representations like attention maps, because the differences between categories may depend on the relations between objects, rather than their mere presence.

Another important task factor is the time available for inspecting the images. We imposed no time constraints, just like most previous studies on human-CNN similarity (Hwu et al., 2021; Karargyris et al., 2021; Lanfredi et al., 2021; Muddamsetty et al., 2021; Trokielewicz et al., 2019; Yang et al., 2022). However, this created a large variance in fixation counts as viewing times ranged from less than 1000 ms up to our data inclusion threshold of around 7000 ms. To test how time constraints affected our results, we repeated our analyses with attention maps that only included the first fixation. The impacts on the results were negligible. Overall, overlaps with Grad-CAM were almost identical between maps derived from all fixations versus only the first fixation (spontaneous gaze: .24 vs .26, gaze-pointing: .26 vs. .25). Also in the interaction with image type we only observed minor changes. Thus, future studies could constrain viewing time for efficiency purposes.

## Particularities of images, image categories, and image types

Our images were selected with the aim of retaining the variability of real-world scenes, instead of choosing a highly homogenous set of prototypical exemplars. Obviously, this increased the variance in our results. However, there also were non-intended but systematic sources of variance. Recall that each image type consisted of two categories selected for their similarity. While analyzing our data, we noticed that these categories were not always as similar as we had intended them to be. This was most noticeable in the case of indoor scenes: for dining rooms, the results resembled those for objects (as attention was focused on the table), while for offices, the results resembled those for landscapes (as attention was distributed across the room). In fact, human-CNN similarity in the most attended areas for dining rooms did not differ from lighthouses and windmills but was higher than for deserts and wheat fields. The opposite was true for offices, with human-CNN similarity being lower than for lighthouses and windmills and as low as for deserts and wheat fields. Interestingly, this stronger focus for dining rooms was not reflected in the total area sizes of drawings, which were similar to those for landscapes. In that sense, wheat fields were an exception, with the total areas being smaller than for deserts, but also smaller than for offices and dining rooms.

These observations imply that future studies should go beyond a simple differentiation of image types based on their superordinate category (e.g., indoor scene). But how to select theoretically interesting and practically relevant image types? First, such differentiation could be based on computations of purely physical image features, such as a scene's distribution of spatial frequencies (Torralba & Oliva, 2003) or its complexity and clutter (Wu et al., 2014). Alternatively, differentiations could be based on human ratings, for instance of a scene's global properties (Greene & Oliva, 2009), the diagnosticity of objects (Wiesmann & Võ, 2023), or scene function (Greene et al., 2016). Finally, a third option is to differentiate scenes based on factors that are known to make classification difficult for CNN: complex and varied scenes with different scales, multiple target objects, varied perspectives, partial occlusions, and untypical or variable lighting conditions (cf. Beery et al., 2018; Mohseni et al., 2021). Future research should investigate which differentiations between image types exert most influence on human-CNN similarity.

## Limitations of the present study

While the present study generated some interesting insights, several limitations should be considered. Limitations concerning our tasks and image types have already been discussed in the previous sections. Moreover, we need to consider methodological limitations of our stimulus material, the experimental design, the data analysis procedures, as well as conceptual limitations.

A first limitation concerns the generalizability of the present findings given our selection of stimuli. We only used six categories and a small set of similar image exemplars within each category. For instance, our objects were long vertical buildings, usually embedded in a natural scene, which calls into question whether the results would generalize to other objects in other contexts. However, it should be noted that our reasoning has never been about all objects merely by virtue of being an object. A close-up view of flower petals spanning the entire image would also be considered an object. However, such an image would actually be more similar to our landscapes in that it would consist of large uniform areas. Accordingly, the attention maps would also have been less focused than they were for our objects. Therefore, our three image types should not be taken too literal, but should be interpreted in the function they were actually meant to fulfil: differentiating between images that can be categorized on the basis of either a single and spatially restricted diagnostic object, object-to-object relations, or global properties. Within these functionally defined image types, future studies should test whether the present results generalize to a larger set of different exemplars and categories. Such generalization attempts should explicitly vary image context, for instance by including object images that present the singular relevant object embedded in a cluttered room. In situations like this, the task-relevant object is already targeted by the first or second fixation (Henderson et al., 2009), due to powerful scene guidance. However, given that eye movements are also attracted by anchor objects (Boettcher et al., 2018) and semantically related objects (Hwang et al., 2011), it will be interesting how such context variations affect attention maps and thus human-CNN similarity.

Second, it should be considered that due to our fixed task order, the three tasks were not independent from each other. We chose this approach on purpose for three reasons. First, we wanted to keep spontaneous gaze unconditional on viewing history, second, we wanted to minimize the need to scan the image during gaze-pointing, and third, we were not concerned that task order would affect drawing. But can this repeated viewing account for some of our results, for instance that the eye movement tasks produced highly similar attention maps? Did participants simply revisit the areas they had previously identified? Indeed, studies that repeatedly exposed participants to the same scenes reported some similarities in their scanpaths (Harding & Bloj, 2010; Underwood et al., 2009). At the same time, the absolute similarity in these studies was quite low, although it reliably differed from chance. Other studies presented the same scenes several times and found that with increasing familiarity, people performed fewer and longer fixations, shorter saccades, and spent less time on meaningful areas (Lancry-Dayan et al., 2019). This might suggest that our methodological choice to always place gaze-pointing second is not without risks, because it confounds task effects with preview and practice effects. We are more optimistic that the fixed task order had no or only minimal impacts on drawing. For one, manual selection is a highly intentional task, and for another, participants could take as much time as they wanted to plan their drawing (and additional analyses of pre-drawing image viewing times indicated that they did). Presumably, it does not make much of a difference whether familarization with the image is spread across three instances of viewing it versus one instance of viewing it for longer. Still, the present study does not allow us to test this assumption, thus future studies could explicitly control for task order effects, either by systematically varying it or by manipulating the tasks between participants.

Third, our methods of computing and comparing attention maps deserve a critical evaluation. One issue is that we only weighted fixations by their duration instead of normalizing them for each participant. Both approaches have their costs and benefits, and thus we also repeated our analyses with normalized data. However, this did not have any remarkable impacts on the results: the overall overlap with Grad-CAM only increased from .24 to .25 for spontaneous gaze, and remained numerically identical for gaze-pointing. Another issue concerns the metrics we used to compare our attention maps. Both the Dice score and cross-correlations have complementary benefits and costs, as discussed above. However, there are problems that neither of them can fix. For instance, they do not control for central fixation bias, do not take the spatial distance between selected areas into account, and do not differentiate between false positives and false negatives. These issues can be addressed by other metrics (Bylinskii et al., 2019). A further issue in comparing our attention maps relates to the statistical method we chose. F2 ANOVAs use images as the degrees of freedom while summing fixations and drawings over all participants. Such summation is the standard procedure performed in all previous human-CNN comparisons. Our statistical comparison goes beyond most previous studies, which typically did not perform inference statistics but merely compared the mean similarity values descriptively (but see Zhang et al., 2019, for an exception that also used F2 ANOVAs). Still, other statistical procedures, such as linear mixed models, would allow for a consideration of both inter-image and inter-human variance. However, this approach would not be applicable here, especially for eye movement maps, because many participants only produced one or a few fixations. Accordingly, their attention maps would be tiny, rendering a comparison with CNN impossible. It probably is for good reason that no previous study used individual participant maps for this purpose. Summing over participants seems unavoidable in this context.

Finally, an important conceptual limitation lies in the very nature of attention maps as a purely spatial method. At best, attention maps can tell us *where* humans or CNN attend, but not *how* they use the information presented in this area (cf. Singh et al., 2020). Thus, despite a high overlap in the attended areas, humans and CNN might attend to completely different features. For instance, CNN rely on object texture more strongly than on shape, while humans do the reverse (Baker et al., 2018; Geirhos et al., 2019). Still, the attention maps of humans and CNN would highlight the same object. One way to consider this in the comparison between human and CNN is to generate different maps for different CNN layers (Ebrahimpour et al., 2019; van Dyck et al., 2021), which process different types of information (Bau et al., 2017). For humans, an indicator of different information processing activities is fixation duration, with longer durations reflecting higher processing depth (Velichkovsky, 2002). Accordingly, attention maps look totally different when they are generated only from focal fixations with long durations and short saccades (high level of processing) than from ambient fixations with short durations and long saccades (low level of processing). Thus, differentiating between different processing levels of humans and CNN seems both desirable and technically feasible.

## Conclusions

What is a suitable task to elicit attention maps for comparisons with CNN? We suggest that when relevant image areas can be located unambiguously, manual selection is an easy way of producing consistent maps that are free from task-irrelevant viewing biases. However, it is less suitable when the relevance of image areas is up to human preferences and may not even be consciously accessible to the human observer. In this case, eye tracking may be a better methodological choice, as the resulting maps are less variable and arbitrary. The specific eye movement task does not matter that much, and thus it seems unnecessary to instruct participants to intentionally fixate relevant areas. Comparing different manual tasks remains an open issue for future studies. To avoid the circular reasoning of inferring the suitability of tasks from the similarity they create, external criteria are needed, such as whether the resulting attention maps are interpretable by humans and CNN (Rong et al., 2021; Zhang et al., 2019).

From a theoretical perspective, future research should strive for a better understanding how humans select the areas they consider relevant for categorization. How much does this depend on the image contents versus on the particular implementation of categorization tasks? How do attention maps depend on individual differences? This is relevant especially in application areas where CNN should not be similar to just any human, but to humans with particular characteristics (e.g., experts). After all, high similarity might also mean that both humans and CNN are mistaken about the actual relevance of image areas (cf. Singh et al., 2020). Thus, it will remain a challenging endeavour how to assess similarities and differences in the information processing of human and artificial agents.

# Acknowledgments

We want to thank Marius Thoß for support in generating the stimuli and Christopher Lindenberg for support in programming the CNN.

# Financial disclosure statement

Initials of authors who received the funding: RM, SS (via heads of chairs: Sebastian Pannasch, Ronald Tetzlaff)

Grant numbers: DFG, PA 1232/15-1; DFG, TE 257/37-1

Full name of each funder: German Research Foundation (Deutsche Forschungsgemeinschaft, DFG)

URL of each funder website: https://www.dfg.de/

The funders had no role in study design, data collection and analysis, decision to publish, or preparation of the manuscript.

# Supporting information

**S1 Fig. Schematic visualization of our ResNet-152 implementation. For full details see the research code in our OSF repository.** Most of the layers are contained in bottleneck blocks inside one of four bottleneck sequences and thus not explicitly shown. Each bottleneck sequence reduces the resolution and increases the number of feature maps. The dashed arrows depict the residual connections (i.e., the addition of the intermediate results from previous layers which are the distinctive feature of the ResNet approach).

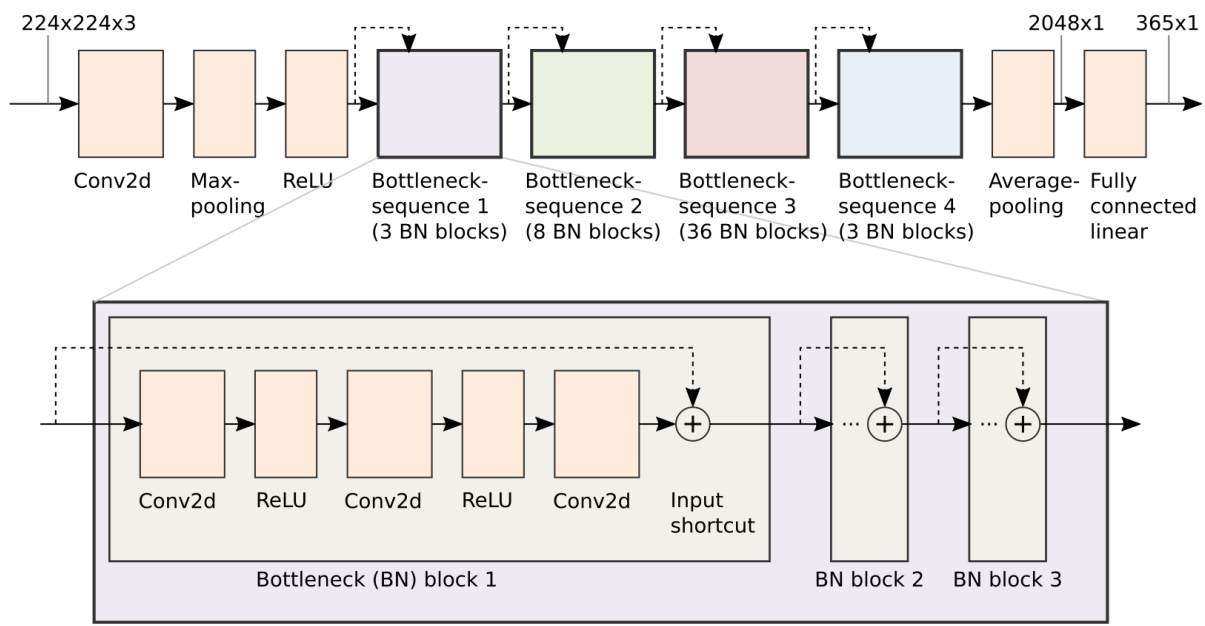

**S1 Table. Details and hyperparameters for the training process of the CNN.**

| Type of detail or hyperparameter | Specification or value |
|---|---|
| Framework | Pytorch |
| Optimizer | SGD |
| Learning rate (LR) | 0.0001 |
| LR decay per epoch | 0.9 |
| Momentum | 0.9 |
| Batch size | 128 |
| Epochs | 10 |
| Random seed | 4427094 |

**S2 Fig. Loss curve and accuracy curves for training and test data.**

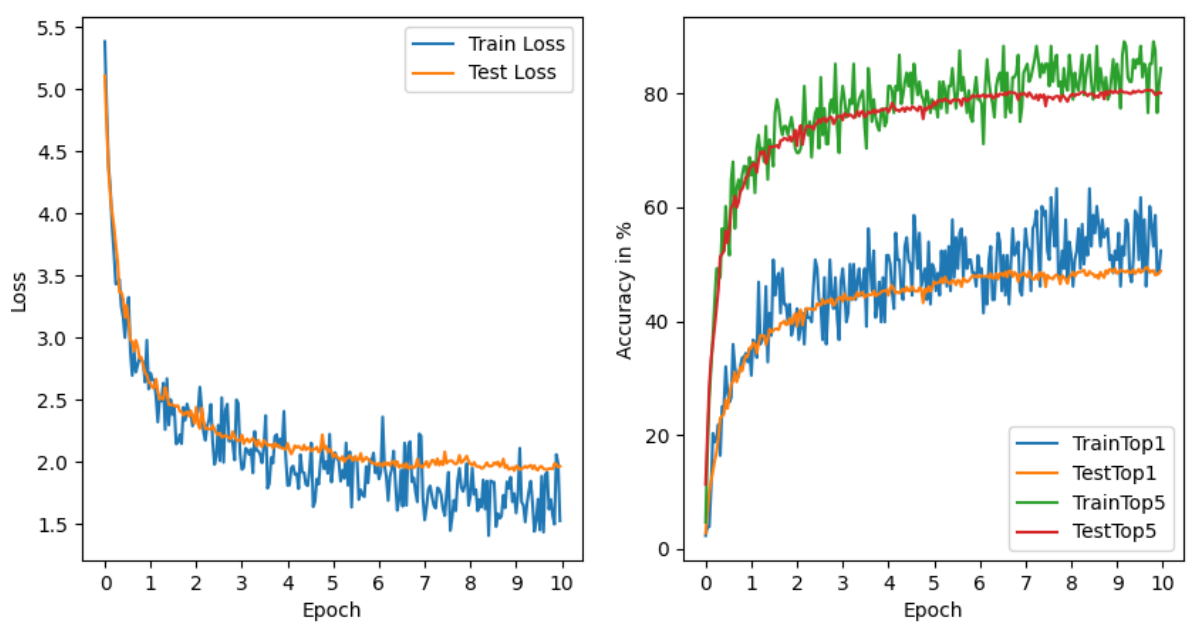